\documentclass[10pt,twocolumn,letterpaper]{article}

\usepackage{iccv}              

\usepackage{xcolor}
\usepackage{amsmath}
\usepackage{amssymb}
\usepackage{mathtools}
\usepackage{amsthm}
\usepackage{bm}
\usepackage{multirow}
\usepackage{subcaption}
\usepackage{booktabs}       
\usepackage{tcolorbox}
\tcbuselibrary{breakable}
\usepackage{graphicx}
\usepackage{enumitem}
\usepackage{caption}
\usepackage{marvosym}
\usepackage{nicematrix}
\usepackage{algorithm}
\usepackage{algorithmic}

\usepackage{listings}

\usepackage{colortbl}
\definecolor{nmgray}{RGB}{229,229,229}
\theoremstyle{plain}

\theoremstyle{definition}

\theoremstyle{remark}

\usepackage[textsize=tiny]{todonotes}

\definecolor{iccvblue}{rgb}{0.21,0.49,0.74}
\usepackage[pagebackref,breaklinks,colorlinks,allcolors=iccvblue]{hyperref}

\newcommand{\ours}[0]{{$\mathrm{T}^2$-VLM}\xspace}

\def\paperID{12475} 
\def\confName{ICCV}
\def\confYear{2025}

\title{Training-free Generation of Temporally Consistent Rewards from VLMs}

\author{%
Yinuo Zhao\textsuperscript{1,$\diamond$}\quad
Jiale Yuan\textsuperscript{3}\quad
Zhiyuan Xu\textsuperscript{2}\quad
Xiaoshuai Hao\textsuperscript{4}\quad
Xinyi Zhang\textsuperscript{1}\quad
Kun Wu\textsuperscript{2}\quad\\
Zhengping Che\textsuperscript{2}\quad
Chi Harold Liu\textsuperscript{1}\textsuperscript{$\dagger$}\quad
Jian Tang\textsuperscript{2}\textsuperscript{$\dagger$}\\[4pt]
\normalsize{\textsuperscript{1}Beijing Institute of Technology}\quad
\normalsize{\textsuperscript{2}Beijing Innovation Center of Humanoid Robotics}\quad\\
\normalsize{\textsuperscript{3}Taobao \& Tmall Group of Alibaba}\quad
\normalsize{\textsuperscript{4}Beijing Academy of Artificial Intelligence}\quad
\\
\tt\small yinuozhao007@gmail.com, \{eric.xu,gongda.wu,z.che,jian.tang\}@x-humanoid.com
}

\begin{document}
\maketitle
\footnotetext{\textsuperscript{$\dagger$} Corresponding authors: Jian Tang (jian.tang@x-humanoid.com) and Chi Harold Liu (liuchi02@gmail.com). \textsuperscript{$\diamond$} This work is done during Yinuo Zhao's internship at Beijing Innovation Center of Humanoid Robotics.}

\begin{abstract}
Recent advances in vision-language models (VLMs) have significantly improved performance in embodied tasks such as goal decomposition and visual comprehension.
However, providing accurate rewards for robotic manipulation without fine-tuning VLMs remains challenging due to the absence of domain-specific robotic knowledge in pre-trained datasets and high computational costs that hinder real-time applicability.
To address this, we propose \ours, a novel training-free, temporally consistent framework that generates accurate rewards through tracking the status changes in VLM-derived subgoals.
Specifically, our method first queries the VLM to establish spatially aware subgoals and an initial completion estimate 
before each round of interaction.
We then employ a Bayesian tracking algorithm to update the goal completion status dynamically, using subgoal hidden states to generate structured rewards for reinforcement learning (RL) agents. 
This approach enhances long-horizon decision-making and improves failure recovery capabilities with RL.
Extensive experiments indicate that \ours achieves state-of-the-art performance in two robot manipulation benchmarks, demonstrating superior reward accuracy with reduced computation consumption.
We believe our approach not only advances reward generation techniques but also contributes to the broader field of embodied AI.
Project website:~\href{https://t2-vlm.github.io/}{https://t2-vlm.github.io/}
\end{abstract}
\begin{figure}[tb]
    \centering
    \subfloat{\includegraphics[width=0.950\columnwidth]{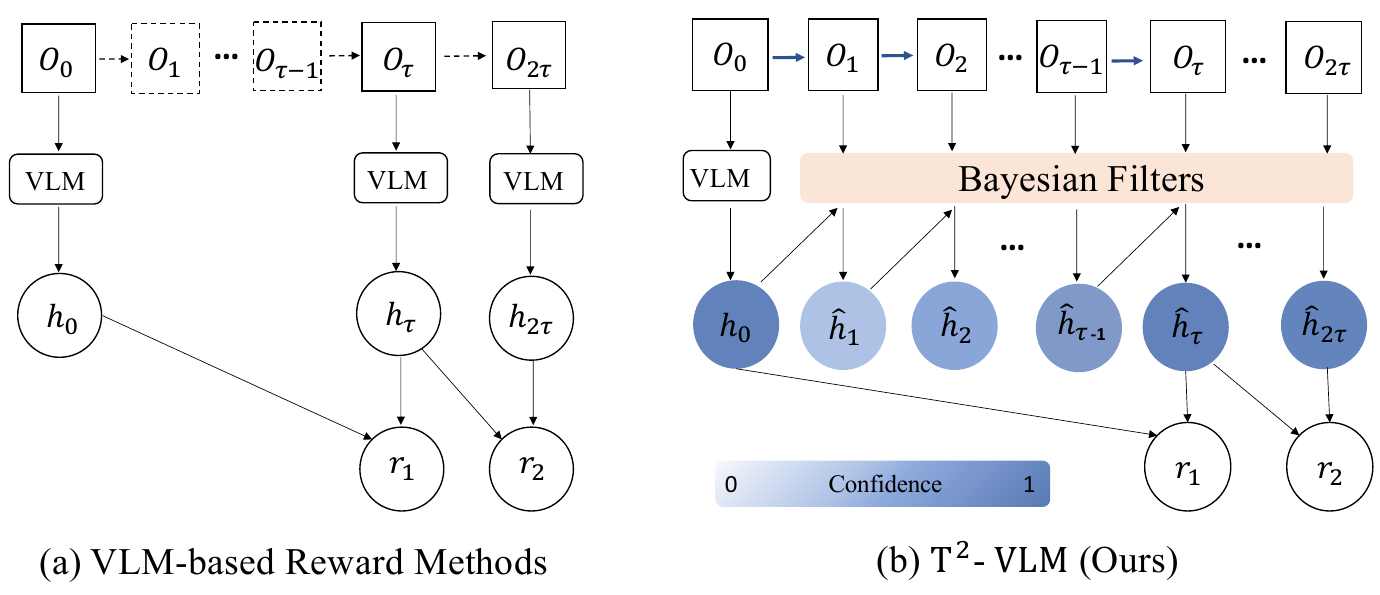}}
    \caption{
 \textbf{Illustrative example of current VLM-based reward generation methods and \textbf{\ours}.}
$o$ represents high-dimensional observations, such as images. $\tau$ denotes the robot's decision intervals. $\bm{h}$ and $\hat{\bm{h}}$ indicate the goal completion status given by VLMs and Bayesian filters, respectively. $r$ is the final reward calculated based on changes in $\bm{h}$. \ours~only requires querying the VLMs once at the beginning of an episode and efficiently generates accurate rewards using Bayesian filters.
    }
    \label{fig:intuition}
\vspace{-1em}
\end{figure}

\section{Introduction}
Pretrained foundation models, such as vision-language models (VLMs)~\cite{wang2024qwen2,radford2021learning,li2023blip,jia2021scaling} and foundation vision models (FVMs)~\cite{kirillov2023segment,ravi2024sam}, exhibit remarkable generalization capabilities across various multimodal tasks~\cite{hao2023mixgen,wu2024robomind} through large-scale pretraining. Recently, they have been widely adopted to advance the robot decision-making process in embodied tasks such as task decomposition~\cite{gao2024physically}, vision understanding~\cite{zhangsam,tan2025reason}, and providing feedback on goal completion~\cite{duvideo}. The feedback can be used to generate rewards for evaluating behavior during  RL interactions, enabling RL training with minimal human supervision.

Existing research on querying VLMs for reward generation can be grouped into two main categories. The first group~\cite{venutocode} generates reward functions explicitly with VLMs before robot manipulation. Although reducing computational costs, they are limited to tasks with simple visual information. 
Other research generates reward signals by identifying goal completion status using VLMs, based directly on images and language descriptions~\cite{wangrl,guo2024doremi,duvideo}. They are applied to more complex robot manipulation tasks, but requires large computational costs. Moreover, it is challenging to generate accurate rewards from VLMs with only spatial observations.
Recently, several studies have integrated temporal information of object trajectories into VLMs to develop more robust policies~\cite{li2024foundation,zheng2024tracevla,wen2023any}.
Motivated by their success, we aim to \textbf{\textit{correct VLM's spatial reasoning outputs by incorporating temporal information}}, thereby improving reward accuracy while reducing the computational costs associated with frequent VLM queries.

In this paper, we propose a novel framework, \ours, a \underline{T}raining-free \underline{T}emporal-consistent reward generation based on the goal decomposition results from VLMs.
As shown in Fig.~\ref{fig:intuition}, the key intuition behind our method is to query the VLMs to initialize the subgoals as hidden state and to continuously update the goal completion status based on temporal observations.
To achieve this, we first design an automated procedure to obtain spatial-aware subgoals with an image and a linguistic task description. Then, we propose a Bayesian tracking algorithm to update the subgoal completion status using temporal observations.
While Bayesian filters~\cite{mucke2024deep,lee2020multimodal} are widely used in robotics to estimate environmental states from multi-modal temporal inputs, they struggle to handle high-dimensional data like images.
To address this, we first leverage the powerful unified segmentation model SAM 2~\cite{ravi2024sam} to extract object bounding box trajectories from high-dimensional inputs. Next, we design a VLM-coding affordance module to derive the observation likelihood and update the Bayesian filters according to it. 
Finally, rewards are computed by evaluating the differences in subgoal hidden states between each decision-making step.
We tested \ours on six robot manipulation tasks involving rigid and articulated objects across three domains. 
Extensive comparisons and ablation studies demonstrate that \ours achieves state-of-the-art reward generation performance in long-horizon tasks. 
Furthermore, the RL agent trained with \ours exhibits a higher recovery success rate and fewer recovery meta-steps compared to other off-the-shelf models when facing online failures.

Our main contributions are summarized as follows:
\begin{itemize}
    \item We propose \ours: a novel training-free temporal-consistent framework based on VLMs for generating accurate rewards in robot manipulation tasks.
    
    
    \item To combine temporal high-dimensional observations, we propose a Bayesian tracking algorithm integrated with LLM-coding affordance to update goal completion status initialized by VLMs.
    
    \item 
Extensive experimental results both in simulation and real-world demonstrate that \ours generates accurate rewards for robot manipulation tasks. Moreover, the agent trained with \ours exhibits a higher recovery success rate and fewer recovery meta-steps compared to other off-the-shelf models.

\end{itemize}

\section{Preliminary}
Bayesian filters effectively integrate multi-modal observations with dynamic priors, making them well-suited for object tracking and for estimating environment states when temporal information is involved. The environment state at time $t$ can be inferred using Bayes' theorem:
\begin{equation}
p(s_t \mid o_{1:t}) = \frac{p(o_t \mid s_t) \, p(s_t \mid o_{1:t-1})}{p(o_t \mid o_{1:t-1})},   
\end{equation}\label{eqn:bayes}
where $s$ denotes the environment states (e.g., object spatial information), $o$ represents the observations (e.g., RGB images), and $t$ indicates the time step. Estimating the environment state involves evaluating Eq.~(\ref{eqn:bayes}) using temporal data from multiple sensors.

The particle filter method is a classical approach within the family of Bayes filters, used to estimate the posterior distribution \( p(s_t) \) given a history of observations. It models the posterior distribution of a stochastic process using a set of particles, typically Gaussian samples, informed by partial observations. Each particle is assigned a weight based on the likelihood of the observation given the estimated state produced by that particle. The particles are then resampled according to their weights, with higher-weight particles retained to estimate the posterior in the next time step.

\section{\texorpdfstring{$\mathrm{T}^2$-VLM}{T²-VLM}}
\begin{figure*}[tb]
    \centering
    \subfloat{\includegraphics[width=0.93\textwidth]{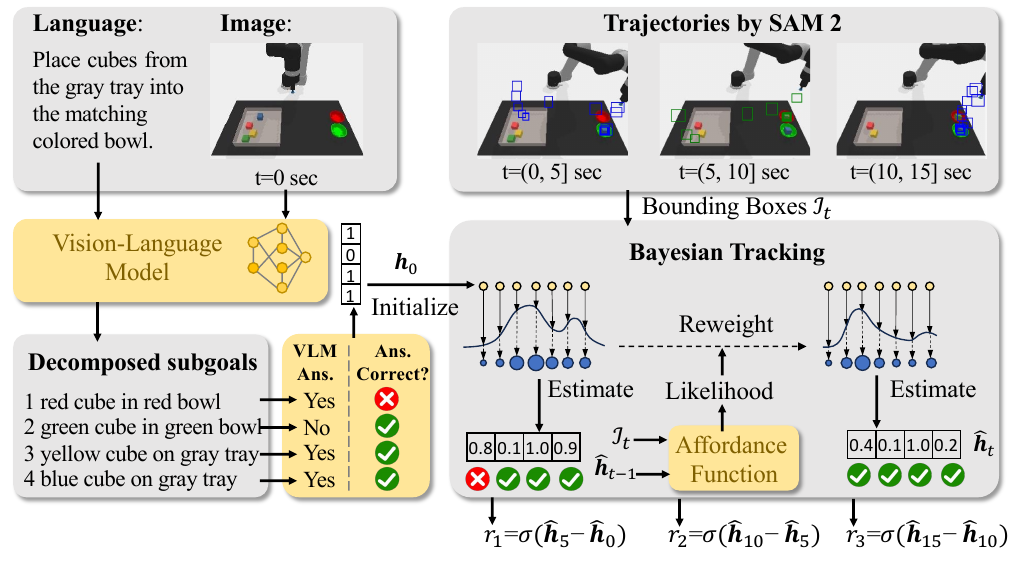}}
    \vspace{-0.3cm}
    \caption{\textbf{An overview of \ours~}. 
    First, the VLM decomposes the subgoals and provides an initial estimate of the goal completion status based on an image and language task description. Then, we introduce a Bayesian tracking algorithm to update the goal completion status using the temporal data processed by SAM 2.
    Finally, the model ensures temporal consistency while accurately generating rewards by tracking the completion status of the environment's goals.
    }
    \label{fig:overview}
    \vspace{-0.6em}
\end{figure*}
In this work, we introduce \ours, a novel reward generation framework that ensures temporal consistency by tracking the environment’s goal completion status.  
Fig.~\ref{fig:overview} provides an overview of \ours.  
Unlike previous VLM-based reward generation methods, \ours queries the VLM only once per episode, then incorporates temporal information (object bounding boxes) to update the VLM-initialized goal status.  
In this section,  we begin by explaining how VLMs are prompted to generate subgoals and initial estimations, then describe a particle filter method that uses VLM-coded affordances for posterior computation, and finally present \ours's implementation details.

\subsection{Prompt VLMs to Generate Subgoals and Provide Initial Estimation}
By leveraging large-scale pretraining, VLMs excel at image captioning and language reasoning, making them a natural choice for automatically decomposing subgoals in long-sequential robotic tasks. As shown in the left part of Fig.~\ref{fig:overview}, the VLM receives an RGB image and a task description as inputs, then generates subgoals step by step. Assuming the initial image captures all task-relevant objects, we design a chain-of-thought image caption prompt to identify task-relevant objects. Next, using the identified objects and the task description, the VLM produces subgoals by analyzing their spatial relationships. Please refer to Fig.~\ref{fig:cot_detection} to Fig.~\ref{fig:goal_generation} in the Appendix for the prompt details. 

Considering large foundation models make more mistakes as the complexity of reasoning tasks increases~\cite{khot2023decomposedpromptingmodularapproach,patel-etal-2022-question}, we decompose tasks with multiple subgoals into simpler ones to better leverage VLMs' zero-shot capabilities. Specifically, we simplify the visual understanding task into a stey-by-step visual question answering (VQA) format, where VLMs are prompted to output ``yes" or ``no" to each subgoal based on the initial RGB image. This VQA process is illustrated in Fig.~\ref{fig:goal_status} and Fig.~\ref{fig:prompts_example} in the Appendix. Despite this design, as shown in the bottom-left of Fig.~\ref{fig:overview}, VLM outputs may still be incorrect due to discrepancies between the large-scale pretraining data and specific robotic scenarios. Our Bayesian tracking algorithm is then designed to incorporate temporal data to correct subgoal completion status from VLM initializations and ensure accurate rewards.

\subsection{Update Subgoal Completion Status with Particle Filters}
As shown in Fig.~\ref{fig:overview}, we represent the initial subgoal completion status as an $N$-dimensional binary vector $\bm{h}_0$. Here, 1 indicates a subgoal is marked as complete by the VLM, and 0 indicates otherwise. We call this vector the subgoal hidden state, reflecting the current spatial relationships among objects in the environment.

Particle filters are classical Bayesian tracking algorithms that used for state estimation and multi-modal fusion. However, they struggle with high-dimensional data without a proper feature extractor. To address this, we use the advanced foundation vision model SAM 2~\cite{ravi2024sam} to track task-related objects in videos and process bounding boxes with the obtained masks. We then employ a Bayesian tracking algorithm to estimate the subgoal hidden state using them.

The particle filters in \ours~estimate the posterior distribution $p(\bm{h}_t|\mathcal{I}_t)$ of the subgoal hidden state $\bm{h}_t$, given the bounding boxes $\mathcal{I}_t$ preprocessed by SAM 2. Each particle represents a hypothesis of the subgoal hidden state. Although subgoal completion status is binary (completed or uncompleted), we constrain particle values to a continuous range between 0 and 1. This continuous representation allows a soft estimation, where higher values indicate greater confidence in subgoal completion at the current timestep.

The tracking process proceeds as follows: First, we initialize all $K$ particles with $\bm{h}_0$ and propagate them using a motion model detailed later. Then, the observation likelihood $p(\bm{h}_t|\mathcal{I}_t)$ for each particle is computed using the observed bounding boxes, and the weights $\omega_k$ are updated accordingly. Systematic resampling, as described in \cite{yang2022particle}, is then employed to select the particles $h_t^{1:K}$. This process involves generating a uniformly distributed random number $u$ from $U[0, 1)$ and computing the following for each particle:
\begin{equation}\label{eqn:reweight1}
    u^k = \frac{k-1 + u}{K},
\end{equation}
\begin{equation}\label{eqn:reweight2}
    \bm{h}_t^k = \bm{h}_{t-1}^i, \text{ where } i \text{ satisfies } u^k \in \left[ \sum_{j=1}^{i-1} \omega_j, \sum_{j=1}^i \omega_j \right].
\end{equation}

Next, we describe the two primary components of our particle filters: the motion module and the observation likelihood module.

\paragraph{Motion Module.} 
The motion module is a fundamental component of particle filters that evolves the states of particles from timestep $t-1$ to $t$. Assuming that the latent hidden state changes smoothly with consistent robot manipulation, we adopt a constant velocity motion model following~\cite{yang2022particle} to simulate the propagation of the latent state:
\begin{equation}
p(\bm{h}_t^k|\bm{h}_{t-1}^k, \bm{h}_{t-2}^k) = \mathcal{N}(\mu, \Sigma),    
\end{equation}
where $\mathcal{N}$ is the multivariate normal distribution with mean $\mu=h^k_{t-1}+\alpha(h_{t-1}^k, h_{t-2}^k)$ and covariance matrix $\Sigma$. $\alpha$ is a soft update hyperparameter that controls the update velocity. A higher value of $\alpha$ enhances robustness in hidden state estimation, especially when facing mis-detected or undetected bounding boxes from SAM 2 due to occlusions or knowledge discrepancies. The covariance matrix $\Sigma=\beta^2 \bm{I}$, where $\bm{I}$ is an identify matrix. 
In our implementation, we set $\alpha=0.7$ and $\beta=0.04$ following~\cite{yang2022particle}.

\paragraph{Observation Likelihood with VLM-Coding Affordance.}
The likelihood $p(\mathcal{I}_t \mid \bm{h}_t)$ measures how likely the observed bounding box $\mathcal{I}_t$ is, given the particle state $\bm{h}_t$ at time $t$. Computing this likelihood analytically is often challenging. For example, consider the distribution $p(\mathcal{I}_t \mid \bm{h}_t=0)$, which gives the position distribution of a “red cube” when the first subgoal, “red cube in red block” is unsatisfied. In this case, the position distribution of the red block can be multimodal and highly complex, making an analytical solution intractable. In our method, we leverage the reasoning and coding abilities in VLMs to solve this challenge. 

\begin{figure}[tb]
    \centering
    \subfloat{\includegraphics[width=0.95\columnwidth]{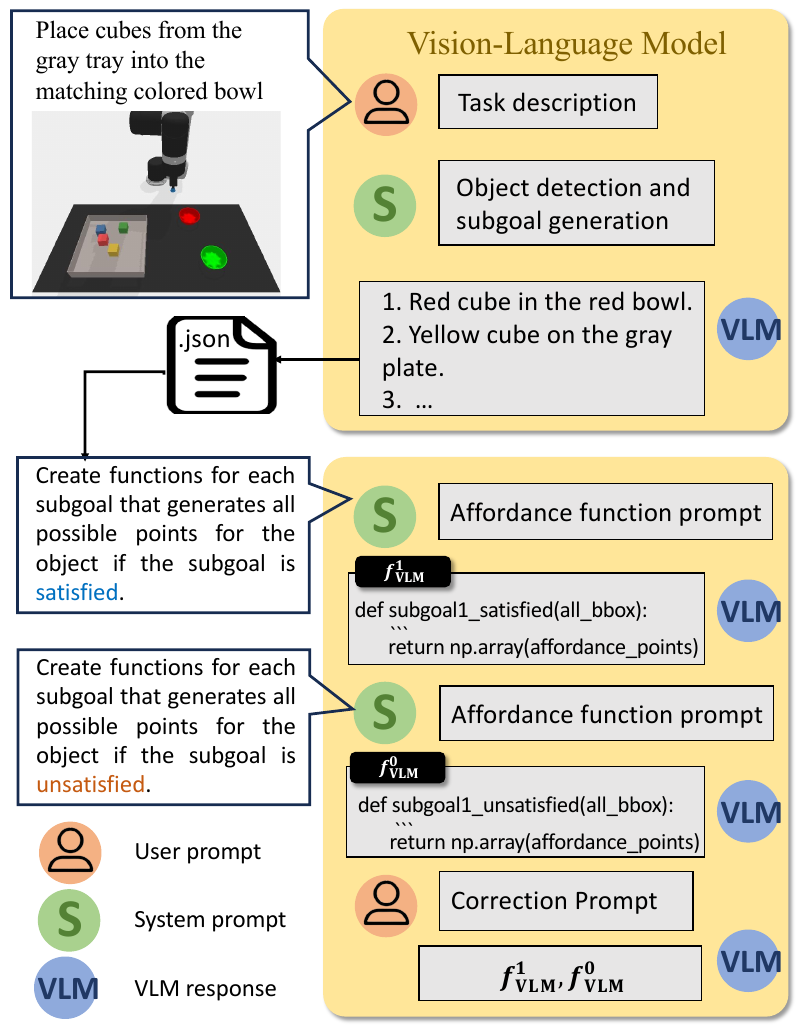}}
    \caption{VLM-coding affordance in observation models.}
    \label{fig:coding_affordance}
    \vspace{-0.5em}
\end{figure}

As shown in Fig.~\ref{fig:coding_affordance}, we query VLMs for possible object positions that either satisfy or do not satisfy the goal. Rather than directly generating these positions, we query VLMs to produce coding functions, which is more efficient and requires no extra fine-tuning. Specifically, for each subgoal $i$, we query VLMs for a function $f^1_{VLM}(i)$ that generates all possible positions of the target object when the $i$-th subgoal is satisfied. The output affordance is denoted by $\hat{\mathcal{I}}^1(i) = \{(x_j, y_j)\}$ (omitting subscript $t$ for brevity). Then, referring to $f^1_{VLM}(i)$, we query VLMs for a function $f^0_{VLM}(i)$ that outputs positions when the $i$-th subgoal is unsatisfied, denoted by $\hat{\mathcal{I}}^0(i) = \{(x_j, y_j)\}$. We then compute the likelihood as the minimum mean distance between the observed positions and the affordances as:
\begin{equation}\label{eqn:distance_function}
    d^0_i = \min_{j \in |\hat{\mathcal{I}}^0|} \bigl\| \mathcal{I} - \hat{\mathcal{I}}^0_j \bigr\|\,, \quad
    d^1_i = \min_{j \in |\hat{\mathcal{I}}^1|} \bigl\| \mathcal{I} - \hat{\mathcal{I}}^1_j \bigr\|\,,
\end{equation}
where $d^0_i$ is the estimation error when subgoal $i$ is unsatisfied ($h_i=0$), and $d^1_i$ is the estimation error when subgoal $i$ is satisfied ($h_i=1$).
We compute the weight $\omega_k$ for each particle $k$ by multiplying the corresponding error terms across all subgoals:
\begin{equation}\label{eqn:weight}
    \omega_k = \prod_{i=1}^N \bigl(h^k_i d^0_i + (1 - h^k_i) d^1_i\bigr),
\end{equation}
where $N$ is the number of subgoals, and $h^k_i$ is the hidden state of particle $k$ for subgoal $i$. This formulation assigns higher weights to particles whose subgoal states match the observed positions (\textit{i.e.}, closer to the satisfied affordance). Finally, we normalize the weights to sum to one and use them to resample particles for the next timestep.

\begin{figure*}[tb]
    \centering
    \subfloat{\includegraphics[width=1\textwidth]{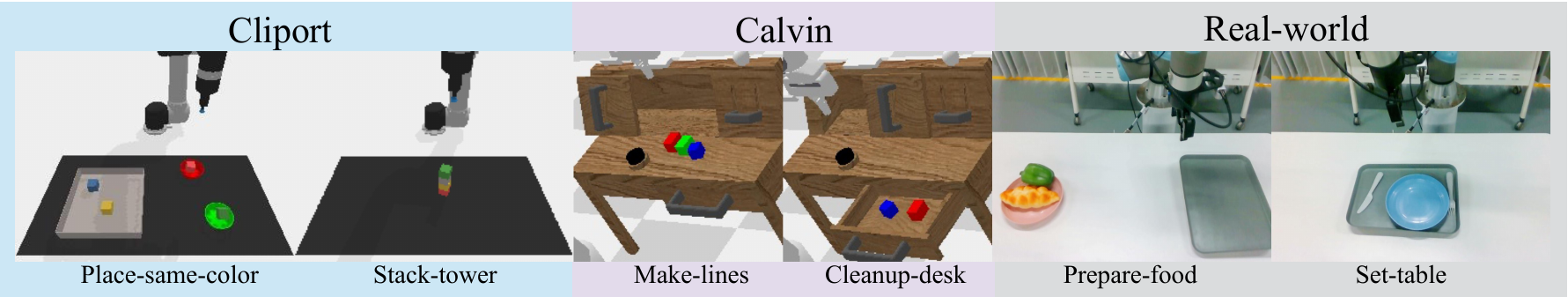}}
    \caption{\textbf{Experiments across three domains with six distinct tasks.} In CLIPort~\cite{shridhar2022cliport}, tasks are ``Place-same-color" where cubes are sorted into matching color bowls, and ``Stack-tower" where cubes are stacked in a predefined color order. In CALVIN~\cite{mees2022calvin},  tasks are ``Make-line" where cubes are aligned in a line following a specific color order, and ``Cleanup-desk" requiring placing cubes into a closed drawer. In real-world, the tasks include ``Prepare-food" where items are arranged into plate, and ``Set-table" where tableware is placed appropriately.  }
    \label{fig:env-settings}
    \vspace{-0.5em}
\end{figure*}

After estimating the subgoal hidden state, we can generate rewards based on changes in goal completion between decision intervals. Specifically, the estimated goal completion status is the weighted average of all particles, \textit{i.e.}, $\hat{\bm{h}}_t = \sum_k \omega_k \bm{h}_t^k$. Then, we compute the reward as $r_t = \sigma(\hat{\bm{h}}_t - \hat{\bm{h}}_{t - \tau})$, where $\sigma(\cdot)$ depends on the decomposed subgoals for the downstream task, and $\tau$ denotes the decision interval. The full detailed procedure of reward generation in \ours~can be found in Algorithm~\ref{algo}.

\subsection{Implementation Details}

In \ours, we adopt Qwen2-VL-72B~\cite{wang2024qwen2} as the default VLM and also experiment with other VLMs. For particle filters, we follow \cite{yang2022particle} with 100 particles, $\alpha=0.7$, and $\beta=0.04$. For SAM2, the official \texttt{sam2.1\_hiera\_tiny} checkpoint provides satisfactory bounding boxes. We define $\sigma$ as a weighted sum of the absolute differences between subgoal hidden states. In our reinforcement-learning setup, we use an object-centric observation and action space~\cite{ma2024explorllm} with a transformer-based network. 
For details of the network architectures and hyper-parameters utilized in the soft-actor-critic RL agent, please refer to the Appendix.


\begin{algorithm}[t!]
\caption{Particle filters in \ours} \label{algo}
\begin{algorithmic}[1]
\footnotesize
\INPUT Bounding box trajectories $\mathcal{I}_{t=1:T}$ of all objects in one episode, initial subgoal hidden state $\bm{h}_0$ from VLM, decision interval $\tau$.
\OUTPUT Rewards for each decision step
\STATE Initialize the estimated hidden state buffer $\mathcal{H}$ = \{$h_0$\}.
\FOR{$t=1$ to $T$}
    \FOR{ $i=1$ to $N$}
    \STATE Obtain estimated position $\hat{\mathcal{I}}^0(i)$  and $\hat{\mathcal{I}}^1(i)$ from the VLM-generated functions $f^0_{VLM}$ and $f^1_{VLM}$ given the bounding box trajectories $\mathcal{I}_t$.
    \STATE Estimate the observation likelihood through computing the distance $d_i^0$ and $d_i^1$ in Eq.~(\ref{eqn:distance_function}).
    \ENDFOR 
    \STATE Compute $\omega_k$ for each particle $k$ through Eq.~(\ref{eqn:weight}).
    \STATE  $\omega_{1:K} \leftarrow \frac{\omega_{1:K}}{\sum\omega_{1:K}}$ 
    \STATE  $\hat{\bm{h}}_t = \sum_k \omega_k \bm{h}_t^k$
    \STATE Insert $\hat{\bm{h}}_t$ into $\mathcal{H}$
    \STATE Resample $\bm{h}_t^{1:K}$ according to Eq.~(\ref{eqn:reweight1}), Eq.~(\ref{eqn:reweight2}).
    \IF{$t$ is decision time step} \item $r_t = \sigma(\hat{\bm{h}}_t - \hat{\bm{h}}_{t - \tau})$ \ENDIF
    \ENDFOR 
\end{algorithmic}
\end{algorithm}

\section{Experiments}
In this section, we address four research questions:
1) \textit{Is the reward generated by \ours~sufficient to support RL training?}
2) \textit{How do \ours-trained policies compare with other methods in terms of recovery capabilities?}
3) \textit{How does each component contribute to \ours?}
4) \textit{Which key factors influence the performance of \ours?}
We investigate these questions through experiments in three domains with six manipulation tasks (see Fig.~\ref{fig:env-settings}).

\begin{figure*}[!ht]
    \centering
    \includegraphics[width=0.7\textwidth]{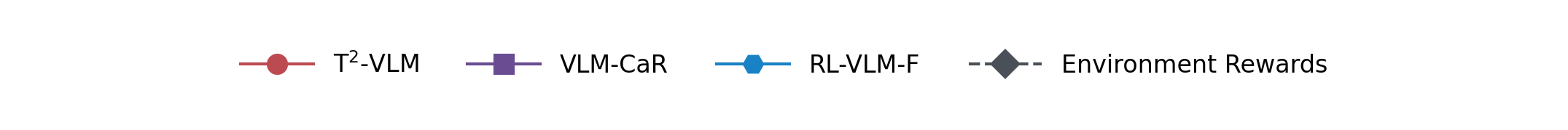}
    \vspace{-0.3cm}  
    
    \subfloat[Place-same-color]{\includegraphics[width=0.23\textwidth]{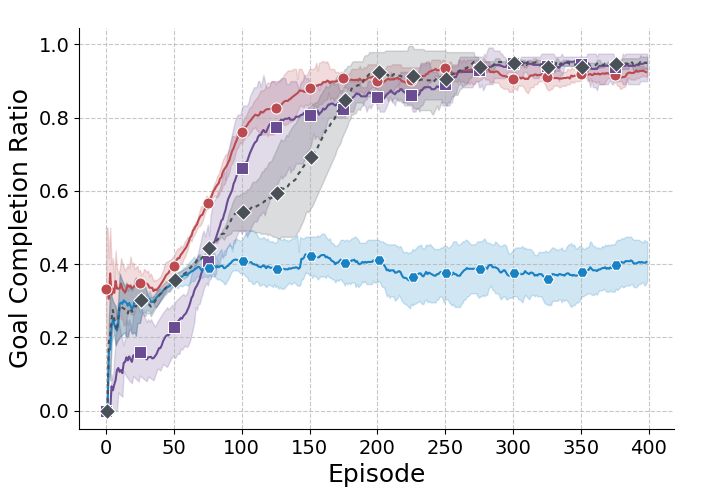}\label{fig:training1}}
    \subfloat[Stack-tower]{\includegraphics[width=0.23\textwidth]{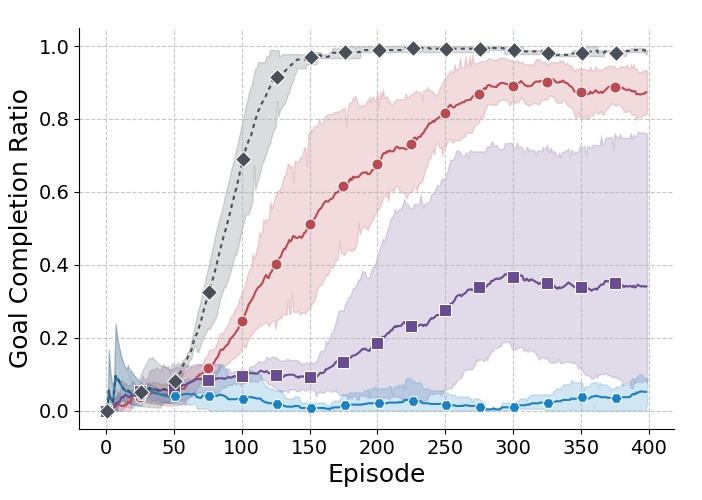}\label{fig:training2}}
    \subfloat[Cleanup-desk]{\includegraphics[width=0.23\textwidth]{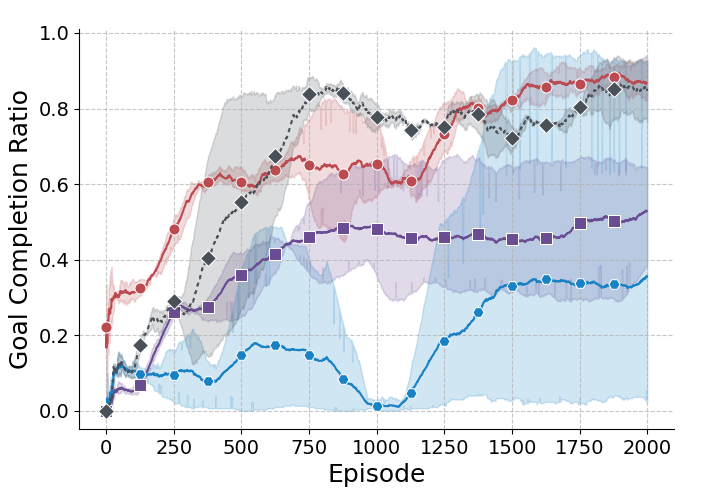}\label{fig:training3}}
    \subfloat[Make-line]{\includegraphics[width=0.23\textwidth]{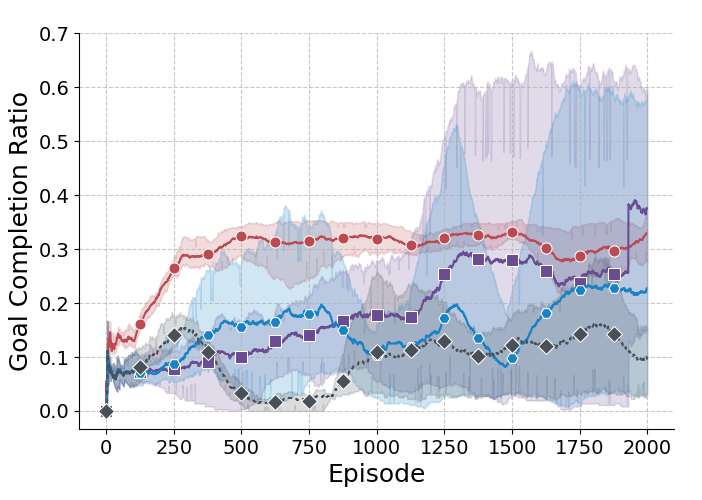}\label{fig:training4}}
    
    \caption{\textbf{Goal completion ratio during RL training under different VLM-based rewards.} All methods are running with three seeds.}
    \label{fig:training}
\end{figure*}

\subsection{Comparison on RL Training}
\textbf{Experiment Setup.}
To answer Question 1, we first introduce four robot manipulation tasks from two commonly used benchmark CLIPort~\cite{shridhar2022cliport} and CALVIN~\cite{mees2022calvin}. They are all long-horizon tasks that involve multiple subgoals. As shown in Fig.~\ref{fig:env-settings}, the success conditions for each task are:
\begin{enumerate}
    \item Place-same-color: The red cube is in the red bowl; The green cube is in the green bowl; The yellow cube and blue cube are on the tray.
    \item Stack-tower: Stack the blocks according to the sequence red, yellow, gray and green. 
    \item Make-line: The red cube is next to the left of the green cube; The green cube is next to the left of the blue cube.
    \item Cleanup-desk: The drawer is open; The red cube is on the drawer; The blue cube is on the drawer.
\end{enumerate}

\textbf{Compared Methods.}
To provide rewards for online RL training, we compare \ours~with two other state-of-the-art (SOTA) VLM-driven reward generation methods, as well as with environmental rewards. The first method, \textbf{VLM-CaR}~\cite{venutocode}, queries VLMs to generate code from images and task descriptions, then refines that code using a small set of expert and random-policy trajectories (five each in our experiments). The second method, \textbf{RL-VLM-F}~\cite{wangrl}, learns a reward model from VLM-generated preference labels on pairs of observation images and task descriptions. Finally, the \textbf{Environmental Rewards} are coded within the benchmark using ground truth environment states. 

\textbf{Evaluation Metric.}  
We measure learning performance in long-sequential tasks using the \textit{goal completion ratio}, defined as the number of achieved goals in an episode divided by the total number of subgoals in the task.

\textbf{Experimental Results.}
As shown in Fig.~\ref{fig:training}, we plot the goal completion ratio during RL training over three seeds per method. We observe that rewards generated by \ours~effectively support RL for robot manipulation, yielding a goal completion ratio comparable to that achieved with environmental rewards and outperforming the other two SOTA methods. Furthermore, \ours~provides more stable training performance, contributing to its precise rewards. In contrast, RL-VLM-F shows poor performance on CLIPort and higher variance on CALVIN, likely due to the discrete action spaces complicating VLM-based preferences and the added instability of a learned reward model. In contrast, by not fully relying on VLM outputs and instead using particle filters for correction, \ours~produces reliable rewards in a training-free manner.

\subsection{Comparison of Recovery Performance}
To address Question 2, we compare the RL policy trained by \ours~against other off-the-shelf methods designed for robot manipulation recovery. We do not include the CaR- and RL-VLM-F-trained policies due to their lower training performance.

We introduce failures during evaluation, including:  
1) Occupied failure (in Place-same-color): Other objects occupy the placement space.  
2) Closed failure (in Cleanup-desk): The drawer is randomly closed during execution.  
3) Picking failure (in Stack-tower): The robot fails to pick up the object.  
4) Placement failure (in Stack-tower): The tower falls because of excessive placement force.

\textbf{Compared Methods.} 
1) \textbf{SayCan}~\cite{ahn2022can} produces context-aware language skills. After executing initial plans, we re-prompt SayCan at a low temperature to carry out subsequent instructions until episode ends.
2) \textbf{REFLECT}~\cite{liu2023reflect} integrates LLM reasoning with visual foundation models for failure analysis and corrective actions.

\textbf{Evaluation Metrics.} We measure failure recovery with two metrics:  
1) \textit{Recovery success rate}: Whether the task is successfully completed within a given time limit after a failure.  
2) \textit{Recovery meta steps}: The number of steps required to achieve the final success.  
All methods can only execute at most 20 steps of primitive skills, because we empirically observe that none of them can recover beyond that point.

\textbf{Experimental Results.} 
Tab.~\ref{tab:recovery} shows that RL with \ours~achieves a higher recovery success rate compared to LLM-only approaches. For instance, in the Stack-tower task with a placement failure, our method achieves full recovery, resulting in a 75\% improvement in success rate over REFLECT. This improvement stems from \ours~leveraging temporal information to provide accurate rewards, even under object occlusion, enabling robust RL policies. In contrast, REFLECT relies on single-image inputs for recovery decisions, often leading to incomplete information. Additionally, our method requires fewer meta steps to recover and complete the task. While REFLECT achieves high success rates under occupied failures, our approach consistently uses fewer meta steps, benefiting from the online training mechanism supported by \ours~rewards.

\begin{table}[t]
    \centering
    \resizebox{1.0\columnwidth}{!}{
        \begin{NiceTabular}{c|r|cccc|c}
        \toprule

        &&Occupied & Closed & Drop & Placement& Average\\
        \midrule
        Recover& \textbf{SayCan} &0.80 &0.10 &0.00& 0.00 & 0.23\\
        success& \textbf{REFLECT} &0.95&0.50&0.40&0.25&0.53 \\
        \rowcolor[HTML]{E0F4FF}rate $\uparrow$& \textbf{RL w/ \ours}& \textbf{1.00}&\textbf{0.80}&\textbf{0.90}&\textbf{1.00}&\textbf{0.93}\\

        \midrule
        Recover& \textbf{SayCan} &14.5 &19.6&20.0&20.0&18.53\\
        meta& \textbf{REFLECT} &16.1&16.6&17.1&18.3&17.03\\
        \rowcolor[HTML]{E0F4FF}steps& \textbf{RL w/ \ours}& \textbf{8.0}&\textbf{10.4}&\textbf{10.2}&\textbf{10.3}&\textbf{9.73}\\
        $\downarrow$& \textbf{Optimal} & 6.0 &6.0 & 8.0&10.0& 7.50 \\
        
        \bottomrule

        \end{NiceTabular}
    }
    \caption{\textbf{Comparison of recovery performance facing with different types of failures.} We ignore the ``failure" term for clarity.}
\label{tab:recovery}
\vspace{-0.5em}
\end{table}

\begin{table*}
    \centering
    \resizebox{\textwidth}{!}{
        \begin{tabular}{r|ccc|ccc}
        \toprule
        Task & \multicolumn{3}{c|}{Place-same-color} & \multicolumn{3}{c}{Stack-tower} \\
        & Query Times & Reward Acc. (\%) & Inference Time (s) & Query Times & Reward Acc. (\%) & Inference Time (s) \\
        \midrule
        VLM-score~(Claude-3.5-sonnet) & 872 & 54.77 & 173.45 & 1620 & 75.05 & 220.62 \\
        \rowcolor[HTML]{E0F4FF}\textbf{\ours~(Claude-3.5-sonnet)} & \textbf{200} & \textbf{90.56} & \textbf{48.64} & \textbf{200} & \textbf{84.45} & \textbf{45.10} \\
        $\Delta$ (\%) & 
        \textcolor[rgb]{0,0.5,0}{ $\downarrow$ 77.06} &  
        \textcolor[rgb]{0,0.5,0}{$\uparrow$ 35.79 } & 
        \textcolor[rgb]{0,0.5,0}{$\downarrow$ 71.9} & 
        \textcolor[rgb]{0,0.5,0}{$\downarrow$ 87.65} & 
        \textcolor[rgb]{0,0.5,0}{$\uparrow$ 9.40} & 
        \textcolor[rgb]{0,0.5,0}{$\downarrow$ 79.56} \\ 
        \midrule
        VLM-score~(Qwen2-VL-72B) & 872 & 68.83 & 36.30 & 1620 & 68.12 & 35.27 \\
        \rowcolor[HTML]{E0F4FF}\textbf{\ours~(Qwen2-VL-72B)} & \textbf{200} & \textbf{94.60} & \textbf{30.81} & \textbf{200} & \textbf{82.68} & \textbf{23.13} \\
        $\Delta$ (\%) & 
        \textcolor[rgb]{0,0.5,0}{$\downarrow$ 77.06} &  
        \textcolor[rgb]{0,0.5,0}{$\uparrow$ 25.77} & 
        \textcolor[rgb]{0,0.5,0}{$\downarrow$ 15.12} & 
        \textcolor[rgb]{0,0.5,0}{$\downarrow$ 87.65} & 
        \textcolor[rgb]{0,0.5,0}{$\uparrow$ 14.56} & 
        \textcolor[rgb]{0,0.5,0}{$\downarrow$ 34.42} \\ 
        \midrule
        VLM-score~(Qwen2-VL-7B) & 872 & 42.08 & 14.84 & 1620 & 76.96 & 22.79 \\
        \rowcolor[HTML]{E0F4FF}\textbf{\ours~(Qwen2-VL-7B)} & \textbf{200} & \textbf{83.25} & \textbf{13.02} & \textbf{200} & \textbf{83.45} & \textbf{20.12} \\
        $\Delta$ (\%) & 
        \textcolor[rgb]{0,0.5,0}{$\downarrow$ 77.06} &  
        \textcolor[rgb]{0,0.5,0}{$\uparrow$ 41.17} & 
        \textcolor[rgb]{0,0.5,0}{$\downarrow$ 12.26} &  
        \textcolor[rgb]{0,0.5,0}{$\downarrow$ 87.65} & 
        \textcolor[rgb]{0,0.5,0}{$\uparrow$ 6.49} & 
        \textcolor[rgb]{0,0.5,0}{$\downarrow$ 11.72} \\ 
        \bottomrule
        \end{tabular}

    }
    \caption{\textbf{Investigations on different foundation VLM models utilized in VLM-score and \ours in CLIPort.} Our method is highlighted in blue. Claude 3.5 Sonnet is one of the most widely used closed-source VLMs.  Qwen2-VL-72B is one of the state-of-the-art open-source VLMs, and Qwen2-VL-7B is a variant in Qwen2 with relatively small parameters.}
    \label{tab:vlms}
    \vspace{-0.5em}
\end{table*}

\subsection{Ablation Studies}
To investigate how each component contributes to \ours~(Question 3), we conduct ablation studies on offline datasets collected from the CALVIN and CLIPort benchmarks, each containing 50 mixed expert and random trajectories per task. We measure \textit{reward accuracy}, defined as the ratio of correctly identified subgoal completions.

\begin{figure}[tb]
    \centering
    \subfloat{\includegraphics[width=0.95\columnwidth]{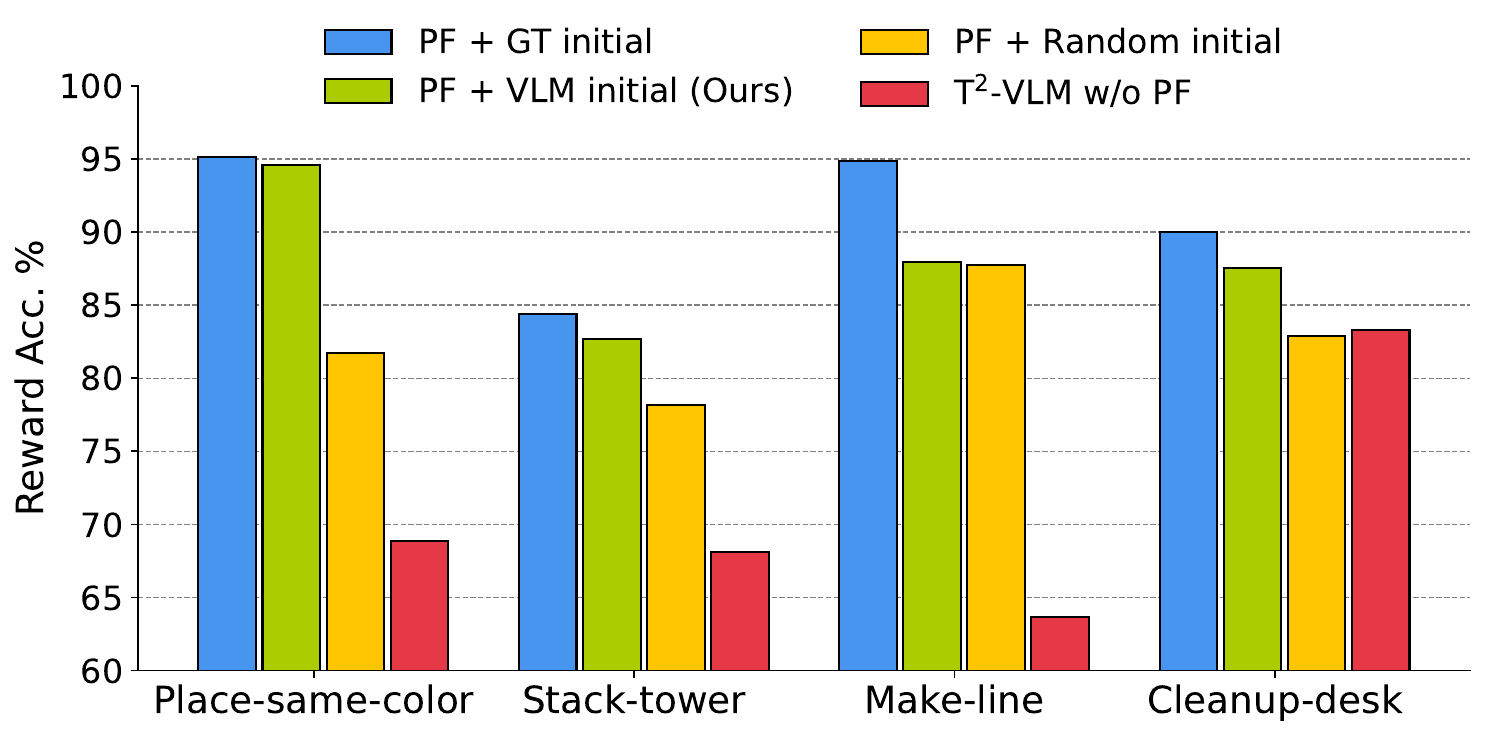}}
    \caption{\textbf{Ablation studies on VLM initialization and particle filters (PF).} We compare the reward accuracy of \ours~under different initialization methods and without particle.}
    \label{fig:ablation}
\vspace{-0.5em}

\end{figure}

As shown in Fig.~\ref{fig:ablation}, we first compare different hidden-state initialization methods for particle filters (PF). Without VLM initialization, PF+Random (yellow bar) achieves only 81.72\% reward accuracy on Place-same-color—12.88\% lower than \ours~(green bar)—highlighting the importance of proper VLM initialization. However, relying solely on VLMs can significantly degrade performance. Removing particle filters and querying VLMs at every step (red bar) reduces reward accuracy below 70\% in the first three tasks due to dynamic conditions (\textit{e.g.}, occlusions). This demonstrates the essential role of particle filters in accurately tracking goal completion over time. The performance of \ours~also depends on task difficulty. In Place-same-color, where VLMs easily identify subgoal completion from the initial image, \ours~achieves reward accuracy comparable to PF with ground-truth initialization (blue bar). However, in tasks like Make-line, where VLMs struggle to recognize the “next to” spatial relationship, \ours~still has room for improvement, suggesting the need for more advanced spatial-reasoning VLMs for initialization.

\subsection{Investigations}
For Question 4, we investigate how different foundation VLMs and PF hidden lengths (the number of images used for reward generation each time) affect the performance. We compare \ours~to VLM-score, which queries VLMs at every step for subgoal completions, focusing on \textit{VLM query times}, \textit{reward accuracy}, and \textit{inference time}. In VLM-score, most inference time comes from frequent VLM queries, whereas in \ours~, it largely stems from bounding box processing (via SAM 2) and particle filter updates.

\textbf{Investigation of Foundation VLMs.} 
We combine \ours~and VLM-score with three different foundation VLMs, as shown in Tab.~\ref{tab:vlms},~\ref{tab:vlms2} and~\ref{tab:vlms3}. We include two real-world tasks here where a topdown camera captures the UR robot workspace. Surprisingly, more powerful VLMs do not always yield better spatial reasoning in robotic scenarios. For example, in the Stack-tower of Tab.~\ref{tab:vlms}, with a longer chain of thought, VLM-score (Claude-3.5-sonnet) achieves around 75\% reward accuracy—on par with VLM-score (Qwen2-VL-7B), while VLM-score (Qwen2-VL-72B) performs even worse despite its larger size. In contrast, \ours~consistently boosts reward accuracy and reduces inference time across all tested VLMs compared with VLM-score. For example, in the Place-same-color task from Tab.~\ref{tab:vlms}, \ours~outperforms VLM-score by 35.79\%, 25.77\%, and 41.17\% in reward accuracy, demonstrating that our method can be integrated with a wide range of VLMs for strong performance. Moreover, \ours~significantly reduces inference time: when combined with Claude-3.5-sonnet for Make-line task in Tab.~\ref{tab:vlms2}, it saves VLM queries by 88.37\%, corresponding to a 50.54\% decrease in total inference time compared with VLM-score.

\begin{table*}
    \centering
    \resizebox{\textwidth}{!}{
        \begin{tabular}{r|ccc|ccc}
        \toprule
        Task  & \multicolumn{3}{c|}{Make-line} & \multicolumn{3}{c}{Cleanup-desk} \\
        &  Query Times & Reward Acc. (\%) & Inference Time (s)  &  Query Times & Reward Acc. (\%) & Inference Time (s)\\
        \midrule
        VLM-score~(Claude-3.5-sonnet)& 860 & 65.43 &166.20&1236&64.36&181.78\\
        \rowcolor[HTML]{E0F4FF} \textbf{\ours~(Claude-3.5-sonnet)}& \textbf{100} & \textbf{89.31} & \textbf{65.59} & \textbf{150} & \textbf{85.07} & \textbf{36.44}\\
        $\Delta$ (\%) & 
        \textcolor[rgb]{0,0.5,0}{$\downarrow$ 88.37} &  
        \textcolor[rgb]{0,0.5,0}{$\uparrow$ 23.88} & 
        \textcolor[rgb]{0,0.5,0}{$\downarrow$ 50.54} & 
        \textcolor[rgb]{0,0.5,0}{$\downarrow$ 87.86} & 
        \textcolor[rgb]{0,0.5,0}{$\uparrow$ 20.71} & 
        \textcolor[rgb]{0,0.5,0}{$\downarrow$ 79.95} \\ 
        \midrule
        
        VLM-score~(Qwen2-VL-72B)& 860 &63.63&42.39&1236&83.33&19.66\\
        \rowcolor[HTML]{E0F4FF} \textbf{\ours~(Qwen2-VL-72B)}& \textbf{100} &\textbf{91.03} & \textbf{42.10} & \textbf{150} & \textbf{93.03}& \textbf{17.14}\\  
        $\Delta$ (\%) & 
        \textcolor[rgb]{0,0.5,0}{$\downarrow$ 88.37} &  
        \textcolor[rgb]{0,0.5,0}{$\uparrow$ 27.40} & 
        \textcolor[rgb]{0,0.5,0}{$\downarrow$ 0.68} & 
        \textcolor[rgb]{0,0.5,0}{$\downarrow$ 87.86} & 
        \textcolor[rgb]{0,0.5,0}{$\uparrow$ 9.70} & 
        \textcolor[rgb]{0,0.5,0}{$\downarrow$ 12.82} \\ 
        \midrule
        VLM-score~(Qwen2-VL-7B)&860 &58.18&39.71&1236&85.93&38.75 \\ 
        \rowcolor[HTML]{E0F4FF} \textbf{\ours~(Qwen2-VL-7B)}& \textbf{100} & \textbf{93.40} & \textbf{24.90} & \textbf{150}& \textbf{92.46} & \textbf{20.17} \\
        $\Delta$ (\%) & 
        \textcolor[rgb]{0,0.5,0}{$\downarrow$ 88.37} &  
        \textcolor[rgb]{0,0.5,0}{$\uparrow$ 35.22} & 
        \textcolor[rgb]{0,0.5,0}{$\downarrow$ 37.30} & 
        \textcolor[rgb]{0,0.5,0}{$\downarrow$ 87.86} & 
        \textcolor[rgb]{0,0.5,0}{$\uparrow$ 6.53} & 
        \textcolor[rgb]{0,0.5,0}{$\downarrow$ 47.95} \\ 
        \bottomrule
        \end{tabular}
    }
    \caption{\textbf{Investigations on different foundation VLM models utilized in VLM-score and \ours in CALVIN.}}
    \label{tab:vlms2}
\end{table*}

\begin{figure}[tb]
    \centering
    \subfloat{\includegraphics[width=0.95\columnwidth]{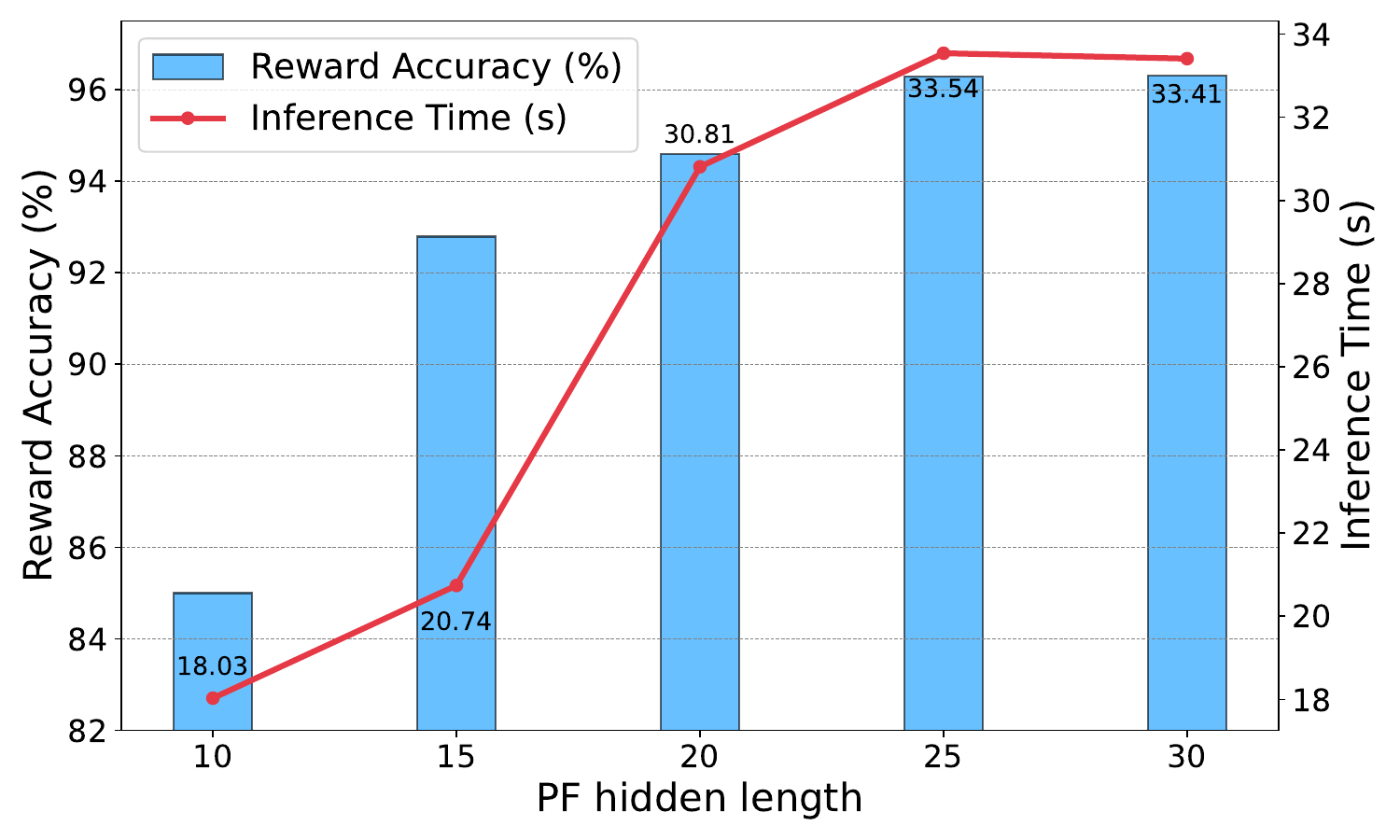}}
    \caption{\textbf{Investigation of PF hidden length.}}
    \label{fig:vary_len}
    \vspace{-0.5em}
\end{figure}

\textbf{Investigation of PF Hidden Length.} Using fewer temporal frames in \ours~can reduce time spent on object tracking but may lower reward accuracy due to less accurate temporal information and insufficient updates for subgoal hidden states. As shown in Fig.~\ref{fig:vary_len}, we vary the PF hidden length in \ours~for the Place-same-color task. Setting the PF hidden length to 25 achieves a good balance between accuracy and efficiency. Beyond that, inference time remains constant because of the fixed sensor frequency.

\section{Related Work}
\textbf{VLM-assisted Robot Manipulation.}
The powerful language generation and generalization capabilities of VLMs have driven significant interest in robotic tasks. One research direction leverages VLMs for spatial reasoning. For example, SORNet~\cite{yuan2022sornet} shows that a transformer architecture conditioned on object prompts can generalize zero-shot to unseen objects. SpatialVLM~\cite{chen2024spatialvlm} predicts spatial relations in metric space; and RoboPoint~\cite{yuanrobopoint} and AffordGrasp~\cite{tang2025affordgrasp} extend spatial reasoning to robotic scenarios. Unlike these methods that directly generate affordance points, \ours~only prompts VLMs to produce functions before task execution, then uses these functions to output potential object positions, which is time-efficient and training-free.
Other works leverage VLMs for long-horizon goal decomposition~\cite{duanmanipulate,liang2023code,singh2023progprompt}, high-level planning~\cite{ji2025robobrain,ahn2022can,liu2023reflect,guo2024doremi,zhang2025mapnav}, and low-level execution~\cite{huang2023voxposer,shridhar2023perceiver}. Our work focus on VLMs’ spatial reasoning capabilities to generate rewards, while we still leverage VLMs to generate subgoals and code to assist in reward generation.

\textbf{VLM-based Reward Generation.}
Recently, VLMs have shown significant potential in generating reward functions, simplifying RL setups. VLM-CaR~\cite{venutocode} uses VLMs to generate reward functions to verify task completion. Similarly, IKER~\cite{patel2024a} samples key points from the scene to create Python-based reward functions using VLMs. However, the generated code often requires human formatting, and may struggle with complex scene interactions or dynamic changes, failing to capture subtle environmental effects. 

Other approaches use VLMs to generate reward values for RL training. Some studies query VLMs for preference labels based on image observations \cite{wangrl,venkataraman2024realworldofflinereinforcementlearning}, while others derive rewards from cosine similarity between environmental states and task descriptions or agent trajectories \cite{rocamonde2024visionlanguage,di2023towards}. VLMs can also output scalar rewards for goal completion given a language-based goal and image observation \cite{baumli2024visionlanguagemodelssourcerewards,yang2023robot}. 
However, they face several challenges including: 
1) Frequent VLM queries incur high computational costs, limiting scalability; 
2) Fine-tuning VLMs requires extensive resources and data, making them impractical for robotics applications; and  
3) Inaccurate reward estimation can lead to suboptimal training outcomes. Therefore, acquiring accurate rewards for robotic tasks    
This motivates the pursuit of more reliable and time-efficient solutions.

\section{Conclusion}

\begin{table}[t]
    \centering
    \resizebox{0.8\columnwidth}{!}{
        \begin{NiceTabular}{r|ccc}
        \toprule

        Task&\multicolumn{2}{c}{Prepare-food}  \\
        \midrule
        &  Reward Acc. (\%) & Inference Time (s)  \\
        Claude-3.5-sonnet & 93.35 & 14.67  \\
        \rowcolor[HTML]{E0F4FF}\textbf{Qwen2-VL-72B} & \textbf{98.56} & \textbf{7.67}\\
        Qwen2-VL-7B &  97.12 & 6.28\\
        \midrule
        Task&\multicolumn{3}{c}{Set-table}  \\
        \midrule
        &  Reward Acc. (\%) & Inference Time (s)  \\
        Claude-3.5-sonnet& 87.61 & 18.68 \\ 
        \rowcolor[HTML]{E0F4FF}\textbf{Qwen2-VL-72B} & \textbf{91.82 }& \textbf{10.59} \\
        Qwen2-VL-7B & 90.36 & 8.64 \\
        \bottomrule

        \end{NiceTabular}
    }
    \caption{\textbf{Investigation on different VLM models combined with \ours~in real world tasks.} \ours~achieves high reward accuracy results when combined with different VLM models.}
\label{tab:vlms3}
\vspace{-0.5em}
\end{table}

Generating reliable and time-efficient rewards remains a major challenge in real-world robot manipulation tasks. In this paper, we propose \textbf{\ours}, a \underline{T}raining-free \underline{T}emporal-consistent reward generation method based on VLM-derived goal decomposition. In this work, we first introduce an automated procedure to prompt VLMs for decomposed subgoals and initial goal completion estimates before interaction, requiring only a single query per episode. Then, we encode these subgoal completion statuses into a scalar vector for particle filter initialization, allowing continuous updates based on temporal observations derived by SAM 2. Experiments across three domains with six robot manipulation tasks show that the \textbf{\ours}~can well support the training of RL algorithms, offering high reward accuracy with lower computational cost.

\section*{Acknowledgments}
This paper was sponsored by the National Key R\&D Project (No. 2023YFE0209100) and National Natural Science Foundation of China (No. U23A20310).

{
    \small
    \bibliographystyle{ieeenat_fullname}
    \bibliography{main}

\begin{thebibliography}{48}
\providecommand{\natexlab}[1]{#1}
\providecommand{\url}[1]{\texttt{#1}}
\expandafter\ifx\csname urlstyle\endcsname\relax
  \providecommand{\doi}[1]{doi: #1}\else
  \providecommand{\doi}{doi: \begingroup \urlstyle{rm}\Url}\fi

\bibitem[Ahn et~al.(2022)Ahn, Brohan, Brown, Chebotar, Cortes, David, Finn, Fu, Gopalakrishnan, Hausman, et~al.]{ahn2022can}
Michael Ahn, Anthony Brohan, Noah Brown, Yevgen Chebotar, Omar Cortes, Byron David, Chelsea Finn, Chuyuan Fu, Keerthana Gopalakrishnan, Karol Hausman, et~al.
\newblock Do as i can, not as i say: Grounding language in robotic affordances.
\newblock \emph{arXiv preprint arXiv:2204.01691}, 2022.

\bibitem[Baumli et~al.(2023)Baumli, Singh, Behbahani, Chan, Comanici, Flennerhag, Gazeau, Holsheimer, Horgan, Laskin, et~al.]{baumli2024visionlanguagemodelssourcerewards}
Kate Baumli, Satinder Singh, Feryal Behbahani, Harris Chan, Gheorghe Comanici, Sebastian Flennerhag, Maxime Gazeau, Kristian Holsheimer, Dan Horgan, Michael Laskin, et~al.
\newblock Vision-language models as a source of rewards.
\newblock In \emph{Second Agent Learning in Open-Endedness Workshop}, 2023.

\bibitem[Chen et~al.(2024)Chen, Xu, Kirmani, Ichter, Sadigh, Guibas, and Xia]{chen2024spatialvlm}
Boyuan Chen, Zhuo Xu, Sean Kirmani, Brain Ichter, Dorsa Sadigh, Leonidas Guibas, and Fei Xia.
\newblock Spatialvlm: Endowing vision-language models with spatial reasoning capabilities.
\newblock In \emph{Proceedings of the IEEE/CVF Conference on Computer Vision and Pattern Recognition}, pages 14455--14465, 2024.

\bibitem[Di~Palo et~al.(2023)Di~Palo, Byravan, Hasenclever, Wulfmeier, Heess, and Riedmiller]{di2023towards}
Norman Di~Palo, Arunkumar Byravan, Leonard Hasenclever, Markus Wulfmeier, Nicolas Heess, and Martin Riedmiller.
\newblock Towards a unified agent with foundation models.
\newblock In \emph{Workshop on Reincarnating Reinforcement Learning at ICLR 2023}, 2023.

\bibitem[Du et~al.(2024)Du, Yang, Florence, Xia, Wahid, Sermanet, Yu, Abbeel, Tenenbaum, Kaelbling, et~al.]{duvideo}
Yilun Du, Sherry Yang, Pete Florence, Fei Xia, Ayzaan Wahid, Pierre Sermanet, Tianhe Yu, Pieter Abbeel, Joshua~B Tenenbaum, Leslie~Pack Kaelbling, et~al.
\newblock Video language planning.
\newblock In \emph{International Conference on Learning Representations}, 2024.

\bibitem[Duan et~al.(2024)Duan, Yuan, Pumacay, Wang, Ehsani, Fox, and Krishna]{duanmanipulate}
Jiafei Duan, Wentao Yuan, Wilbert Pumacay, Yi~Ru Wang, Kiana Ehsani, Dieter Fox, and Ranjay Krishna.
\newblock Manipulate-anything: Automating real-world robots using vision-language models.
\newblock In \emph{8th Annual Conference on Robot Learning}, 2024.

\bibitem[Gao et~al.(2024)Gao, Sarkar, Xia, Xiao, Wu, Ichter, Majumdar, and Sadigh]{gao2024physically}
Jensen Gao, Bidipta Sarkar, Fei Xia, Ted Xiao, Jiajun Wu, Brian Ichter, Anirudha Majumdar, and Dorsa Sadigh.
\newblock Physically grounded vision-language models for robotic manipulation.
\newblock In \emph{IEEE International Conference on Robotics and Automation}, pages 12462--12469, 2024.

\bibitem[Guo et~al.(2024)Guo, Wang, Zha, and Chen]{guo2024doremi}
Yanjiang Guo, Yen-Jen Wang, Lihan Zha, and Jianyu Chen.
\newblock Doremi: Grounding language model by detecting and recovering from plan-execution misalignment.
\newblock In \emph{IEEE/RSJ International Conference on Intelligent Robots and Systems}, pages 12124--12131, 2024.

\bibitem[Hao et~al.(2023)Hao, Zhu, Appalaraju, Zhang, Zhang, Li, and Li]{hao2023mixgen}
Xiaoshuai Hao, Yi Zhu, Srikar Appalaraju, Aston Zhang, Wanqian Zhang, Bo Li, and Mu Li.
\newblock Mixgen: A new multi-modal data augmentation.
\newblock In \emph{Proceedings of the IEEE/CVF winter conference on applications of computer vision}, pages 379--389, 2023.

\bibitem[Huang et~al.(2023)Huang, Wang, Zhang, Li, Wu, and Fei-Fei]{huang2023voxposer}
Wenlong Huang, Chen Wang, Ruohan Zhang, Yunzhu Li, Jiajun Wu, and Li Fei-Fei.
\newblock Voxposer: Composable 3d value maps for robotic manipulation with language models.
\newblock In \emph{Conference on Robot Learning}, pages 540--562, 2023.

\bibitem[Ji et~al.(2025)Ji, Tan, Shi, Hao, Zhang, Zhang, Wang, Zhao, Mu, An, et~al.]{ji2025robobrain}
Yuheng Ji, Huajie Tan, Jiayu Shi, Xiaoshuai Hao, Yuan Zhang, Hengyuan Zhang, Pengwei Wang, Mengdi Zhao, Yao Mu, Pengju An, et~al.
\newblock Robobrain: A unified brain model for robotic manipulation from abstract to concrete.
\newblock In \emph{Proceedings of the Computer Vision and Pattern Recognition Conference}, pages 1724--1734, 2025.

\bibitem[Jia et~al.(2021)Jia, Yang, Xia, Chen, Parekh, Pham, Le, Sung, Li, and Duerig]{jia2021scaling}
Chao Jia, Yinfei Yang, Ye Xia, Yi-Ting Chen, Zarana Parekh, Hieu Pham, Quoc Le, Yun-Hsuan Sung, Zhen Li, and Tom Duerig.
\newblock Scaling up visual and vision-language representation learning with noisy text supervision.
\newblock In \emph{International Conference on Machine Learning}, pages 4904--4916, 2021.

\bibitem[Khot et~al.(2023)Khot, Trivedi, Finlayson, Fu, Richardson, Clark, and Sabharwal]{khot2023decomposedpromptingmodularapproach}
Tushar Khot, Harsh Trivedi, Matthew Finlayson, Yao Fu, Kyle Richardson, Peter Clark, and Ashish Sabharwal.
\newblock Decomposed prompting: A modular approach for solving complex tasks.
\newblock In \emph{The Eleventh International Conference on Learning Representations}, 2023.

\bibitem[Kirillov et~al.(2023)Kirillov, Mintun, Ravi, Mao, Rolland, Gustafson, Xiao, Whitehead, Berg, Lo, et~al.]{kirillov2023segment}
Alexander Kirillov, Eric Mintun, Nikhila Ravi, Hanzi Mao, Chloe Rolland, Laura Gustafson, Tete Xiao, Spencer Whitehead, Alexander~C Berg, Wan-Yen Lo, et~al.
\newblock Segment anything.
\newblock In \emph{Proceedings of the IEEE/CVF International Conference on Computer Vision}, pages 4015--4026, 2023.

\bibitem[Lee et~al.(2020)Lee, Yi, Mart{\'\i}n-Mart{\'\i}n, Savarese, and Bohg]{lee2020multimodal}
Michelle~A Lee, Brent Yi, Roberto Mart{\'\i}n-Mart{\'\i}n, Silvio Savarese, and Jeannette Bohg.
\newblock Multimodal sensor fusion with differentiable filters.
\newblock In \emph{IEEE/RSJ International Conference on Intelligent Robots and Systems}, pages 10444--10451, 2020.

\bibitem[Li et~al.(2024)Li, Jin, Sun, Yu, Shi, Hao, Hao, Liu, Sun, Zhang, et~al.]{li2024foundation}
Dingzhe Li, Yixiang Jin, Yuhao Sun, Hongze Yu, Jun Shi, Xiaoshuai Hao, Peng Hao, Huaping Liu, Fuchun Sun, Jianwei Zhang, et~al.
\newblock What foundation models can bring for robot learning in manipulation: A survey.
\newblock \emph{arXiv preprint arXiv:2404.18201}, 2024.

\bibitem[Li et~al.(2023)Li, Li, Savarese, and Hoi]{li2023blip}
Junnan Li, Dongxu Li, Silvio Savarese, and Steven Hoi.
\newblock Blip-2: Bootstrapping language-image pre-training with frozen image encoders and large language models.
\newblock In \emph{International Conference on Machine Learning}, pages 19730--19742, 2023.

\bibitem[Liang et~al.(2023)Liang, Huang, Xia, Xu, Hausman, Ichter, Florence, and Zeng]{liang2023code}
Jacky Liang, Wenlong Huang, Fei Xia, Peng Xu, Karol Hausman, Brian Ichter, Pete Florence, and Andy Zeng.
\newblock Code as policies: Language model programs for embodied control.
\newblock In \emph{International Conference on Robotics and Automation}, pages 9493--9500, 2023.

\bibitem[Liu et~al.(2023)Liu, Bahety, and Song]{liu2023reflect}
Zeyi Liu, Arpit Bahety, and Shuran Song.
\newblock Reflect: Summarizing robot experiences for failure explanation and correction.
\newblock In \emph{Conference on Robot Learning}, pages 3468--3484, 2023.

\bibitem[Ma et~al.(2024)Ma, Luijkx, Ajanovic, and Kober]{ma2024explorllm}
Runyu Ma, Jelle Luijkx, Zlatan Ajanovic, and Jens Kober.
\newblock Explorllm: Guiding exploration in reinforcement learning with large language models.
\newblock \emph{arXiv preprint arXiv:2403.09583}, 2024.

\bibitem[Mees et~al.(2022)Mees, Hermann, Rosete-Beas, and Burgard]{mees2022calvin}
Oier Mees, Lukas Hermann, Erick Rosete-Beas, and Wolfram Burgard.
\newblock Calvin: A benchmark for language-conditioned policy learning for long-horizon robot manipulation tasks.
\newblock \emph{IEEE Robotics and Automation Letters}, 7\penalty0 (3):\penalty0 7327--7334, 2022.

\bibitem[M{\"u}cke et~al.(2024)M{\"u}cke, Boht{\'e}, and Oosterlee]{mucke2024deep}
Nikolaj~T M{\"u}cke, Sander~M Boht{\'e}, and Cornelis~W Oosterlee.
\newblock The deep latent space particle filter for real-time data assimilation with uncertainty quantification.
\newblock \emph{Scientific Reports}, 14\penalty0 (1):\penalty0 19447, 2024.

\bibitem[Patel et~al.(2022)Patel, Mishra, Parmar, and Baral]{patel-etal-2022-question}
Pruthvi Patel, Swaroop Mishra, Mihir Parmar, and Chitta Baral.
\newblock Is a question decomposition unit all we need?
\newblock In \emph{Proceedings of the 2022 Conference on Empirical Methods in Natural Language Processing}, pages 4553--4569, Abu Dhabi, United Arab Emirates, 2022. Association for Computational Linguistics.

\bibitem[Patel et~al.(2024)Patel, Yin, Huang, Garg, Nayyeri, Fei-Fei, Lazebnik, and Li]{patel2024a}
Shivansh Patel, Xinchen Yin, Wenlong Huang, Shubham Garg, Hooshang Nayyeri, Li Fei-Fei, Svetlana Lazebnik, and Yunzhu Li.
\newblock A real-to-sim-to-real approach to robotic manipulation with {VLM}-generated iterative keypoint rewards.
\newblock In \emph{2nd CoRL Workshop on Learning Effective Abstractions for Planning}, 2024.

\bibitem[Radford et~al.(2021)Radford, Kim, Hallacy, Ramesh, Goh, Agarwal, Sastry, Askell, Mishkin, Clark, et~al.]{radford2021learning}
Alec Radford, Jong~Wook Kim, Chris Hallacy, Aditya Ramesh, Gabriel Goh, Sandhini Agarwal, Girish Sastry, Amanda Askell, Pamela Mishkin, Jack Clark, et~al.
\newblock Learning transferable visual models from natural language supervision.
\newblock In \emph{International Conference on Machine Learning}, pages 8748--8763, 2021.

\bibitem[Ravi et~al.(2024)Ravi, Gabeur, Hu, Hu, Ryali, Ma, Khedr, R{\"a}dle, Rolland, Gustafson, et~al.]{ravi2024sam}
Nikhila Ravi, Valentin Gabeur, Yuan-Ting Hu, Ronghang Hu, Chaitanya Ryali, Tengyu Ma, Haitham Khedr, Roman R{\"a}dle, Chloe Rolland, Laura Gustafson, et~al.
\newblock Sam 2: Segment anything in images and videos.
\newblock \emph{arXiv preprint arXiv:2408.00714}, 2024.

\bibitem[Rocamonde et~al.(2024)Rocamonde, Montesinos, Nava, Perez, and Lindner]{rocamonde2024visionlanguage}
Juan Rocamonde, Victoriano Montesinos, Elvis Nava, Ethan Perez, and David Lindner.
\newblock Vision-language models are zero-shot reward models for reinforcement learning.
\newblock In \emph{International Conference on Learning Representations}, 2024.

\bibitem[Sharan et~al.(2024)Sharan, Zhao, ufuk topcu, Wang, and Chinchali]{sharan2024plan}
S~P Sharan, Ruihan Zhao, ufuk topcu, Zhangyang Wang, and Sandeep~P. Chinchali.
\newblock Plan diffuser: Grounding {LLM} planners with diffusion models for robotic manipulation.
\newblock In \emph{Bridging the Gap between Cognitive Science and Robot Learning in the Real World: Progresses and New Directions}, 2024.

\bibitem[Shridhar et~al.(2022)Shridhar, Manuelli, and Fox]{shridhar2022cliport}
Mohit Shridhar, Lucas Manuelli, and Dieter Fox.
\newblock Cliport: What and where pathways for robotic manipulation.
\newblock In \emph{Conference on robot learning}, pages 894--906, 2022.

\bibitem[Shridhar et~al.(2023)Shridhar, Manuelli, and Fox]{shridhar2023perceiver}
Mohit Shridhar, Lucas Manuelli, and Dieter Fox.
\newblock Perceiver-actor: A multi-task transformer for robotic manipulation.
\newblock In \emph{Conference on Robot Learning}, pages 785--799, 2023.

\bibitem[Singh et~al.(2023)Singh, Blukis, Mousavian, Goyal, Xu, Tremblay, Fox, Thomason, and Garg]{singh2023progprompt}
Ishika Singh, Valts Blukis, Arsalan Mousavian, Ankit Goyal, Danfei Xu, Jonathan Tremblay, Dieter Fox, Jesse Thomason, and Animesh Garg.
\newblock Progprompt: Generating situated robot task plans using large language models.
\newblock In \emph{International Conference on Robotics and Automation}, pages 11523--11530, 2023.

\bibitem[Tan et~al.(2025)Tan, Ji, Hao, Lin, Wang, Wang, and Zhang]{tan2025reason}
Huajie Tan, Yuheng Ji, Xiaoshuai Hao, Minglan Lin, Pengwei Wang, Zhongyuan Wang, and Shanghang Zhang.
\newblock Reason-rft: Reinforcement fine-tuning for visual reasoning.
\newblock \emph{arXiv preprint arXiv:2503.20752}, 2025.

\bibitem[Tang et~al.(2023)Tang, Zheng, Yu, and Yang]{tang2023cotdetaffordanceknowledgeprompting}
Jiajin Tang, Ge Zheng, Jingyi Yu, and Sibei Yang.
\newblock Cotdet: Affordance knowledge prompting for task driven object detection.
\newblock In \emph{Proceedings of the IEEE/CVF International Conference on Computer Vision}, pages 3068--3078, 2023.

\bibitem[Tang et~al.(2025)Tang, Zhang, Hao, Wang, Wu, Wang, and Zhang]{tang2025affordgrasp}
Yingbo Tang, Shuaike Zhang, Xiaoshuai Hao, Pengwei Wang, Jianlong Wu, Zhongyuan Wang, and Shanghang Zhang.
\newblock Affordgrasp: In-context affordance reasoning for open-vocabulary task-oriented grasping in clutter.
\newblock \emph{arXiv preprint arXiv:2503.00778}, 2025.

\bibitem[Venkataraman et~al.(2024)Venkataraman, Wang, Wang, Erickson, and Held]{venkataraman2024realworldofflinereinforcementlearning}
Sreyas Venkataraman, Yufei Wang, Ziyu Wang, Zackory Erickson, and David Held.
\newblock Real-world offline reinforcement learning from vision language model feedback.
\newblock \emph{arXiv preprint arXiv:2411.05273}, 2024.

\bibitem[Venuto et~al.(2024)Venuto, Islam, Klissarov, Precup, Yang, and Anand]{venutocode}
David Venuto, Mohammad Sami~Nur Islam, Martin Klissarov, Doina Precup, Sherry Yang, and Ankit Anand.
\newblock Code as reward: Empowering reinforcement learning with vlms.
\newblock In \emph{International Conference on Machine Learning}, 2024.

\bibitem[Wang et~al.(2024{\natexlab{a}})Wang, Bai, Tan, Wang, Fan, Bai, Chen, Liu, Wang, Ge, et~al.]{wang2024qwen2}
Peng Wang, Shuai Bai, Sinan Tan, Shijie Wang, Zhihao Fan, Jinze Bai, Keqin Chen, Xuejing Liu, Jialin Wang, Wenbin Ge, et~al.
\newblock Qwen2-vl: Enhancing vision-language model's perception of the world at any resolution.
\newblock \emph{arXiv preprint arXiv:2409.12191}, 2024{\natexlab{a}}.

\bibitem[Wang et~al.(2024{\natexlab{b}})Wang, Sun, Zhang, Xian, Biyik, Held, and Erickson]{wangrl}
Yufei Wang, Zhanyi Sun, Jesse Zhang, Zhou Xian, Erdem Biyik, David Held, and Zackory Erickson.
\newblock Rl-vlm-f: Reinforcement learning from vision language foundation model feedback.
\newblock In \emph{International Conference on Machine Learning}, 2024{\natexlab{b}}.

\bibitem[Wen et~al.(2023)Wen, Lin, So, Chen, Dou, Gao, and Abbeel]{wen2023any}
Chuan Wen, Xingyu Lin, John So, Kai Chen, Qi Dou, Yang Gao, and Pieter Abbeel.
\newblock Any-point trajectory modeling for policy learning.
\newblock \emph{arXiv preprint arXiv:2401.00025}, 2023.

\bibitem[Wu et~al.(2024)Wu, Hou, Liu, Che, Ju, Yang, Li, Zhao, Xu, Yang, et~al.]{wu2024robomind}
Kun Wu, Chengkai Hou, Jiaming Liu, Zhengping Che, Xiaozhu Ju, Zhuqin Yang, Meng Li, Yinuo Zhao, Zhiyuan Xu, Guang Yang, et~al.
\newblock Robomind: Benchmark on multi-embodiment intelligence normative data for robot manipulation.
\newblock \emph{arXiv preprint arXiv:2412.13877}, 2024.

\bibitem[Yang et~al.(2023)Yang, Zhang, Li, Zou, Li, and Gao]{yang2023set}
Jianwei Yang, Hao Zhang, Feng Li, Xueyan Zou, Chunyuan Li, and Jianfeng Gao.
\newblock Set-of-mark prompting unleashes extraordinary visual grounding in gpt-4v.
\newblock \emph{arXiv preprint arXiv:2310.11441}, 2023.

\bibitem[Yang et~al.(2024)Yang, Mark, Vu, Sharma, Bohg, and Finn]{yang2023robot}
Jingyun Yang, Max~Sobol Mark, Brandon Vu, Archit Sharma, Jeannette Bohg, and Chelsea Finn.
\newblock Robot fine-tuning made easy: Pre-training rewards and policies for autonomous real-world reinforcement learning.
\newblock In \emph{International Conference on Robotics and Automation}, pages 4804--4811, 2024.

\bibitem[Yang et~al.(2022)Yang, Stork, and Stoyanov]{yang2022particle}
Yuxuan Yang, Johannes~A Stork, and Todor Stoyanov.
\newblock Particle filters in latent space for robust deformable linear object tracking.
\newblock \emph{IEEE Robotics and Automation Letters}, pages 12577--12584, 2022.

\bibitem[Yuan et~al.(2022)Yuan, Paxton, Desingh, and Fox]{yuan2022sornet}
Wentao Yuan, Chris Paxton, Karthik Desingh, and Dieter Fox.
\newblock Sornet: Spatial object-centric representations for sequential manipulation.
\newblock In \emph{Conference on Robot Learning}, pages 148--157, 2022.

\bibitem[Yuan et~al.(2024)Yuan, Duan, Blukis, Pumacay, Krishna, Murali, Mousavian, and Fox]{yuanrobopoint}
Wentao Yuan, Jiafei Duan, Valts Blukis, Wilbert Pumacay, Ranjay Krishna, Adithyavairavan Murali, Arsalan Mousavian, and Dieter Fox.
\newblock Robopoint: A vision-language model for spatial affordance prediction in robotics.
\newblock In \emph{8th Annual Conference on Robot Learning}, 2024.

\bibitem[Zhang et~al.(2024)Zhang, Bai, He, Wang, Zhao, Li, and Li]{zhangsam}
Junjie Zhang, Chenjia Bai, Haoran He, Zhigang Wang, Bin Zhao, Xiu Li, and Xuelong Li.
\newblock Sam-e: Leveraging visual foundation model with sequence imitation for embodied manipulation.
\newblock In \emph{International Conference on Machine Learning}, 2024.

\bibitem[Zhang et~al.(2025)Zhang, Hao, Xu, Zhang, Zhang, Wang, Zhang, Wang, Zhang, and Xu]{zhang2025mapnav}
Lingfeng Zhang, Xiaoshuai Hao, Qinwen Xu, Qiang Zhang, Xinyao Zhang, Pengwei Wang, Jing Zhang, Zhongyuan Wang, Shanghang Zhang, and Renjing Xu.
\newblock Mapnav: A novel memory representation via annotated semantic maps for vlm-based vision-and-language navigation.
\newblock \emph{arXiv preprint arXiv:2502.13451}, 2025.

\bibitem[Zheng et~al.(2024)Zheng, Liang, Huang, Gao, Daum{\'e}~III, Kolobov, Huang, and Yang]{zheng2024tracevla}
Ruijie Zheng, Yongyuan Liang, Shuaiyi Huang, Jianfeng Gao, Hal Daum{\'e}~III, Andrey Kolobov, Furong Huang, and Jianwei Yang.
\newblock Tracevla: Visual trace prompting enhances spatial-temporal awareness for generalist robotic policies.
\newblock \emph{arXiv preprint arXiv:2412.10345}, 2024.

\end{thebibliography}
}
\clearpage
\newpage
\appendix

This supplementary material provides additional details on the proposed method and used prompts that could not be included in the main manuscript due to page limitations.

Specifically, this appendix is organized as follows:
\begin{itemize}
  \item Sec.~\ref{sec:supp_1} provides additional details on the model architecture and reinforcement learning training.
  \item Sec.~\ref{sec:supp_2} presents details on the prompts used for querying VLMs.  
  \item Sec.~\ref{sec:supp_3} presents details for the experimental environment.     
\end{itemize}

\section{Model Architecture and Training Details}
\label{sec:supp_1}

\begin{figure*}[htbp]
    \centering
    \subfloat{\includegraphics[width=0.95\textwidth]{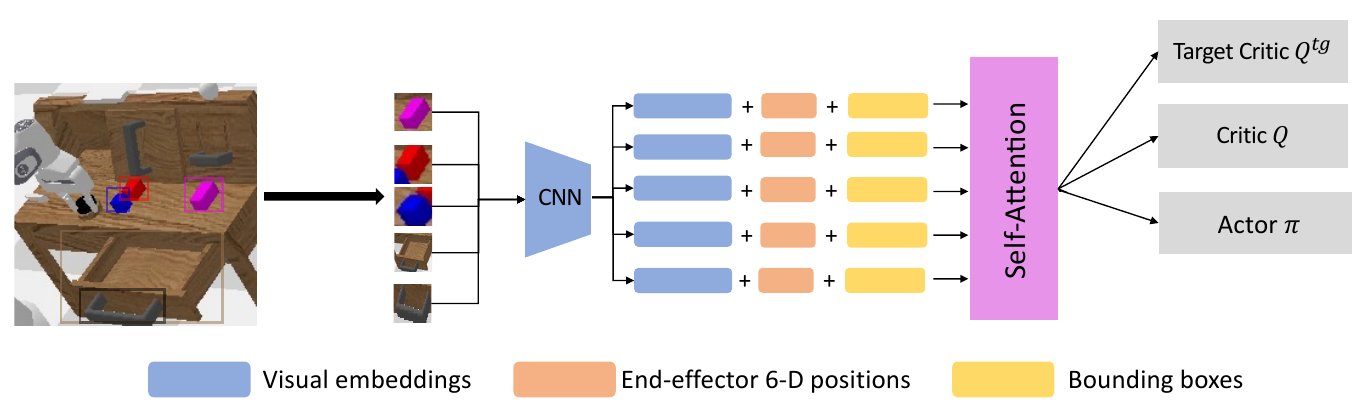}}
    \caption{\textbf{The network architecture for RL agent.}}
    \label{fig:network}
\end{figure*}

We use an object-centric transformer-based network architecture in our RL agent, similar to the one used in ExploRLLM~\cite{ma2024explorllm}. As illustrated in Fig.~\ref{fig:network}, the visual embeddings, end-effector 6-D positions, and object bounding boxes are concatenated for each task-related object. As we concentrate on reward generation in this work, the bounding boxes used here are obtained from the simulators for all methods, following~\cite{ma2024explorllm}. When computing rewards, we still utilizes SAM2 to obtain the bounding boxes without any supervision. Then, these features are input into the self-attention module to extract the overall features of the current frame. During the training phase, the convolutional neural networks (CNN) and the self-attention network are updated concurrently with the policy and value networks. We set the temperature $\alpha$=0, discount factor $\gamma=0.99$, critic update $\tau=0.01$ in SAC. You can also find these details in our open-source code repository: \href{https://github.com/nuomizai/T2VLM}{https://github.com/nuomizai/T2VLM}

\section{Prompts for Querying VLMs}
\label{sec:supp_2}

\begin{figure*}[htbp]
\begin{tcolorbox}[title=Prompts for chain-of-thought object detection,colback=gray!5,colframe=nmgray!75!black,before skip=1pt, after skip=1pt,width=0.8\textwidth,center]
{\footnotesize
\textbf{User input Task Description:}

Place the cube from gray tray into matching colored bowl; cubes without a matching bowl should stay in place.

\textbf{VLM Input:} 

\begin{center}
\includegraphics[width=0.4\textwidth]{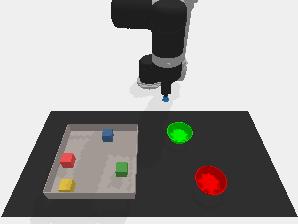}
\end{center}
This is a task related to robot arm manipulation. Pay a comprehensive attention to the provided picture and detect all the objects in detail shown in this image.

\vspace{0.2cm}

\textbf{VLM Output:}

Let's identify all the objects present:

\begin{itemize}
\setlength{\itemsep}{-0.3em}
\item Robot Arm. The black robotic arm is positioned above the gray tray, indicating it is used for picking and placing objects.
\item Gray Tray. The gray tray contains four cubes of different colors:
\begin{itemize}
    \item Red Cube
    \item Blue Cube
    \item Green Cube
    \item Yellow Cube
\end{itemize}
\item Colored Bowls. There are three colored bowls on the table:
\begin{itemize}
    \item Green Bowl: Positioned to the right of the gray tray.
    \item Red Bowl: Positioned further to the right of the green bowl.
    \item No Blue or Yellow Bowls: These colors are missing, meaning there are no bowls for the blue and yellow cubes.
\end{itemize}
\item Table Surface. The objects (gray tray and bowls) are placed on a flat, dark surface which serves as the workspace for the robot arm.
\end{itemize}

\vspace{0.2cm}

\textbf{VLM Input:} 

Assume you are an expert of robot arm control filed with high intelligence. Your goal is to achieve this task: Place the cube from the gray tray into the matching colored bowl; cubes without a matching bowl should stay in place.

Refer to the previous description for the objects shown in image, If the robot arm wants to achieve this task, what task-related objects need to be considered? Please print each object name with specific color like ``white cube". Robot arm could be ignore. Your answer needs to be as detailed as possible to avoid confusion. The output format is a list string:
[
    ``XX",
    ...
]

\vspace{0.2cm}

\textbf{VLM Output:} 

[``a red cube", ``a green cube", ``a blue cube", ``a yellow cube", ``a gray tray", ``a green bowl", ``a red bowl"]

}
\end{tcolorbox}

\vspace{0.2cm}

\caption{\textbf{Prompts for chain-of-thought object detection.}}
\label{fig:cot_detection}
\end{figure*}

\paragraph{Chain-of-thought Task Relevant Object Detection.}
Detecting task-relevant objects poses a significant challenge that requires domain knowledge. Leveraging the strong visual captioning and reasoning abilities of VLM, we generate these objects using chain-of-thought (CoT) prompting, building on previous work~\cite{sharan2024plan, tang2023cotdetaffordanceknowledgeprompting}. Below, we provide an example in Fig.~\ref{fig:cot_detection} illustrating how to identify task-related objects with CoT prompts.

\begin{figure*}[htbp] 
    \centering
    \includegraphics[width=\textwidth]{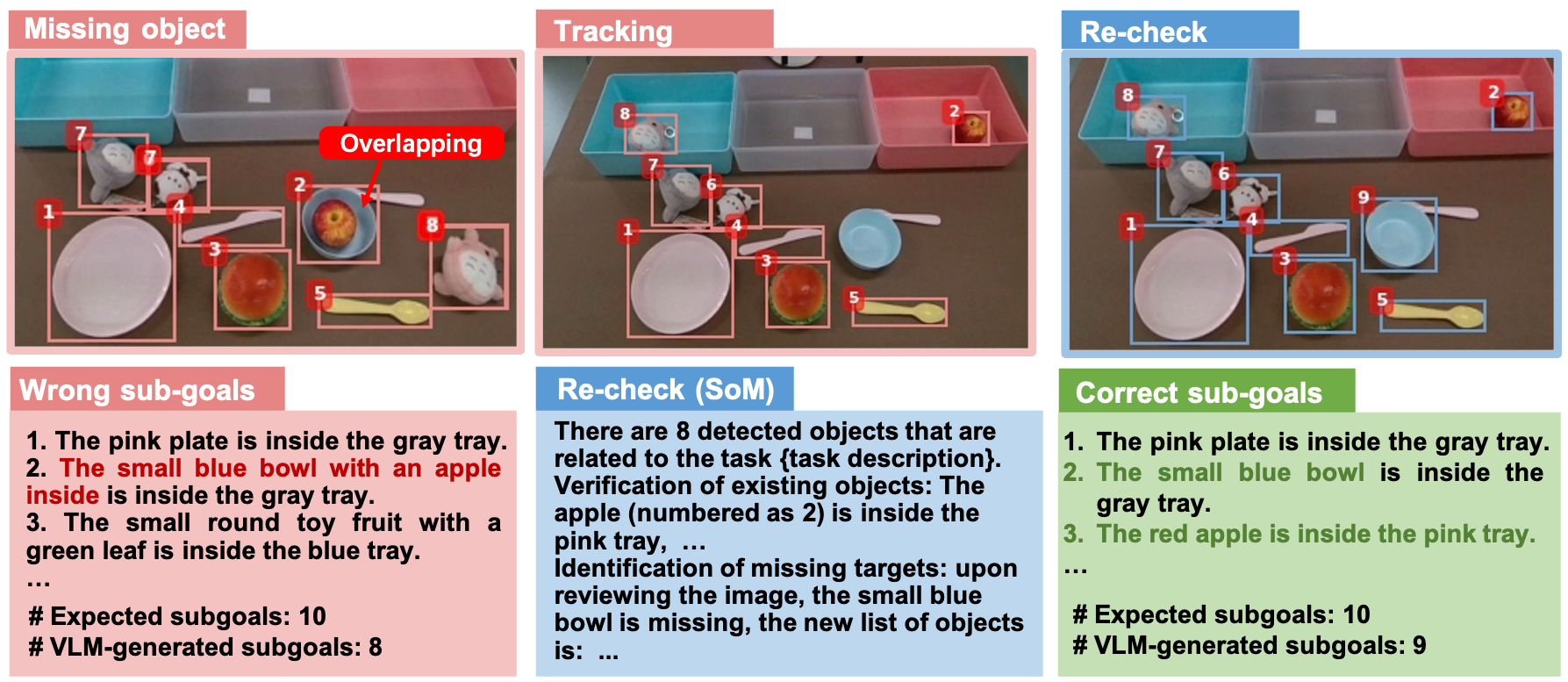}
    \caption{\textbf{Re-check VLM initialization with Set-of-Mark (SoM) prompting in a long-horizon real-world task.}}
    \label{fig:vlm-error}
\end{figure*}

\paragraph{Prompt for Sub-goal Generation.}
As shown in Fig.~\ref{fig:goal_generation}, we provide the prompts for sub-goal generation. Specifically, we input task descriptions specified by the users and task-related object names obtained from Fig.~\ref{fig:cot_detection} into user prompts for sub-goal generation. These prompts allow the VLM to generate spatial-aware sub-goals that can be tracked with Bayesian filters.

Typically, these prompts effectively guide the VLM to generate correct subgoals, provided the initial image captures all task-relevant objects. However, in challenging cases—such as when objects are overlapping or partially hidden—VLMs may fail to recognize key items, leading to false or incomplete subgoals. We find that a simple N-step re-check during initialization can effectively mitigate this issue. As shown in Fig.~\ref{fig:vlm-error}, in a real-world sorting task, the VLM initially merges the apple and blue bowl due to overlap, resulting in missing subgoals. A second query during random exploration—using SoM prompting~\cite{yang2023set} with numbered labels and bounding boxes from \ours—successfully recovers the blue bowl and updates the subgoals. This demonstrates the potential of our method to improve robustness in scenarios where VLMs may overlook relevant objects.

\begin{figure*}[htbp]
\begin{tcolorbox}[breakable, title=Prompts for identifying the completion status of each subgoal, width=0.8\textwidth, colback=gray!5,colframe=nmgray!75!black,before skip=1pt, after skip=1pt,fontupper=\linespread{1.0}\selectfont,center]
{\footnotesize

\textbf{System Prompt:}

\vspace{0.2cm}

You assist in controlling a high-intelligence robotic arm that determine the spatial relationship between the objects in the image. Print Yes or No given the task description and queried subgoal.

\vspace{0.2cm}

\textbf{User Prompt:} 

\vspace{0.2cm}

\emph{Task Description}: Place cubes from the gray tray into the matching colored bowl, cubes without a matching bowl should stay in place.

\begin{center}
\includegraphics[width=0.4\textwidth]{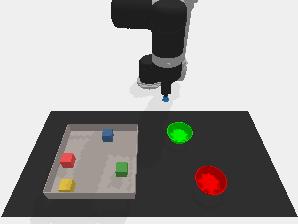}
\end{center}

\emph{Subgoal}: The red cube is in the red bowl.

Please determine if the current status meets the subgoal.

\textbf{VLM output:}  
No.
}

\end{tcolorbox}
\vspace{0.2cm}
\caption{\textbf{Prompts for goal completion status queried VLMs.}}
\label{fig:goal_status}
\end{figure*}

\begin{figure*}[htbp]
\begin{tcolorbox}[title=An example for completion status identification of sub-goals, width=0.8\textwidth, colback=gray!5,colframe=nmgray!75!black,before skip=1pt, after skip=1pt,fontupper=\linespread{1.0}\selectfont,center]

{\footnotesize

\textbf{Task Description:}

\vspace{0.1cm}

Place the cube from the gray tray into the matching colored bowl; cubes without a matching
bowl should stay in place.

\vspace{0.2cm}

\begin{center}
\includegraphics[width=0.4\textwidth]{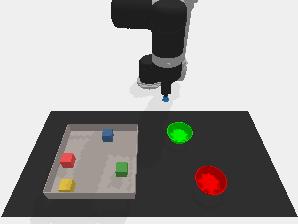}
\end{center}

\textbf{Chain-of-thought Object Detection:}

\vspace{0.1cm}

    blue cube, red cube, green cube, yellow cube, red bowl, green bowl

\vspace{0.1cm}

\textbf{Sub-goal Generation:}

\begin{enumerate}
    \item The blue cube remains on the gray tray.
    \item The red cube is inside the red bowl.
    \item The green cube is inside the green bowl.
    \item The yellow cube remains on the gray tray.
\end{enumerate}

\vspace{0.2cm}

\textbf{Completion Status Identification of Sub-goals:}

\vspace{0.2cm}

\emph{System prompt:} Detailed in Fig.~\ref{fig:goal_status}.

\vspace{0.1cm}

\emph{Task description}: Place the cube from the gray tray into the matching colored bowl; cubes without a matching
bowl should stay in place.

\vspace{0.2cm}

\vspace{0.1cm}

\textbf{Case 1.} Sub-goal matching for \emph{The blue cube remains on the gray tray.} 

\vspace{0.2cm}

\emph{Sub-goal}: The blue cube remains on the gray tray.

\vspace{0.1cm}

Please determine if the current status meets the sub-goal.

\vspace{0.1cm}

\textbf{VLM Output:} 
Yes.

\vspace{0.4cm}

\textbf{Case 2.} Sub-goal matching for: \emph{The red cube is inside the red bowl.}

\vspace{0.2cm}

\emph{Sub-goal}: The red cube is inside the red bowl.

\vspace{0.1cm}

Please determine if the current status meets the sub-goal.

\vspace{0.1cm}

\textbf{VLM Output:} 
No.

\vspace{0.4cm}

\textbf{Case 3.} Sub-goal matching for: \emph{The green cube is inside the green bowl.} 

\vspace{0.2cm}

\emph{Sub-goal}: The green cube is inside the green bowl.

\vspace{0.1cm}

Please determine if the current status meets the sub-goal.

\vspace{0.1cm}

\textbf{VLM Output:} 
No.

\vspace{0.4cm}

\textbf{Case 4.} Sub-goal matching for: \emph{The yellow cube remains on the gray tray.} 

\vspace{0.2cm}

\emph{Sub-goal}: The yellow cube remains on the gray tray.

\vspace{0.1cm}

Please determine if the current status meets the sub-goal.

\vspace{0.1cm}

\textbf{VLM Output:} 
Yes.


}
\end{tcolorbox}\label{box:example}
\vspace{0.2cm}
\caption{\textbf{An example of goal completion status identification.}}
\label{fig:prompts_example}
\end{figure*}

\paragraph{Prompt for Goal Completion Status Identification.}
The prompts for identifying the completion status of each sub-goal are shown in Fig.~\ref{fig:goal_status}. A complete subgoal completion status identification process for the Place-same-color task is visualized in Fig.~\ref{fig:prompts_example}.

\begin{figure*}[htp]
\begin{tcolorbox}[title=Prompts for sub-goals generation, width=0.8\textwidth, colback=gray!5,colframe=nmgray!75!black,before skip=1pt, after skip=1pt,fontupper=\linespread{1.0}\selectfont,center]
{\footnotesize

\textbf{User Input Task Description:}

Place the cube from gray tray into matching colored bowl; cubes without a matching bowl should stay in place.

\vspace{0.2cm}

\textbf{System Prompt:}

\vspace{0.2cm}
You are an intelligent vision-language assistant agent operating within a virtual environment. Your primary objective is to achieve specified goals by breaking down complex tasks into manageable sub-goals that describe the status of objects at various stages.

\vspace{0.1cm}
When you receive a task description, decompose it into a set of clear and concise sub-goals that will enable you to accomplish the task. Ensure that each sub-goal is complete and accurate to prevent task failure due to missing information about the status of relevant objects.

\vspace{0.1cm}
To describe the state of target objects accurately, use the spatial relationship terms like: \emph{inside, on top of, on the bottom of, on the left of, on the right of} and so on. Additionally, use the motion terms like \emph{open, closed} and so on to describe object status. 

\vspace{0.1cm}
Make sure that each description captures the essential intermediate status needed to complete the task.

\vspace{0.2cm}
\textbf{User Prompt:}

\vspace{0.2cm}
Examples:

\vspace{0.1cm}
\emph{Task}: Place the red cube inside the red bowl.

\emph{Relevant Objects}: Red cube, red bowl.

\emph{Sub-Goals}: 1. The red cube is inside the red bowl.

\vspace{0.2cm}
\emph{Task}: Stack three cubes in the order of red, black, and blue, with the red cube on top.

\emph{Relevant Objects}: Red cube, black cube, blue cube.

\emph{Sub-Goals}: 1. The black cube is on top of the blue cube.

2. The red cube is on top of the black cube.

\vspace{0.2cm}
Now, decompose the following task using the provided visual information:

\emph{Task}: Place the cube from the gray tray into the matching colored bowl; cubes without a matching bowl should stay in place.

\emph{Relevant Objects}: a red cube, a green cube, a blue cube, a yellow cube, a gray tray, a green bowl, a red bowl

\emph{Sub-Goals}: 

\vspace{0.2cm}
\textbf{VLM Output:}

\begin{enumerate}
    \item The green cube is inside the green bowl.
    \item The red cube is inside the red bowl.
    \item The blue cube is on top of the gray tray.
    \item  The yellow cube is on top of the gray tray.
\end{enumerate}

}
\end{tcolorbox}
\vspace{0.2cm}
\caption{\textbf{Prompts for sub-goals generation in the setup stage.}}
\label{fig:goal_generation}
\end{figure*}

\section{Experiments}
\label{sec:supp_3}
\subsection{Environment Details}
\paragraph{Training Environments.} To support online RL training, we designed four robot manipulation tasks based on the CLIPort and CALVIN benchmarks. RL algorithms are trained for 400 episodes on CLIPort tasks and 2,000 episodes on CALVIN tasks, due to CALVIN's larger action space and longer execution steps. Training may end early due to task success or unrecoverable failures. In CLIPort, no unrecoverable failures occur. In CALVIN, an episode terminates if the object falls off the desk.

\paragraph{Testing Environments.} For testing the baselines' recovery capability, we modified the training environments by introducing environment changes or robot execution failures. For all methods in Tab.~\ref{tab:recovery}, we measure the average recovery success rate and recovery meta steps over 20 trials. 

\subsection{Action Space}\label{sec:supp_action_space}
In the CLIPort benchmark, following the settings of SayCan~\cite{ahn2022can}, the RL agent needs to simultaneously determine the objects for picking and placing. Therefore, the action space in CLIPort is $\mathcal{A}=[a_1, a_2]$, where $a_1$ is the total number of picking objects and $a_2$ is the total number of placing objects. In the CALVIN benchmark, the action space consists of a list of finite compositions combining skills and objects. Specifically, the action space of each task is shown as follows:
\begin{itemize}
    \item Cleanup-desk task: $\mathcal{A}$=[[PICK, RED], [PICK, BLUE], [PLACE, DRAWER], [PULL, HANDLE], [PUSH, HANDLE]]
    \item Make-line task: $\mathcal{A}$=[[PICK, BLUE], [PLACE2LEFT, BLUE], [PLACE2RIGHT, BLUE], [PICK, RED], [PLACE2LEFT, RED], [PLACE2RIGHT, RED],[PICK, GREEN], [PLACE2LEFT, GREEN], [PLACE2RIGHT, GREEN]]
\end{itemize}

\paragraph{Primitive Skill Parameterization.} The primitive skill considered in simulators is parameterized as SKILL[$\bm{p}$], where $\bm{p}$=[$x$, $y$, $z$] specifies the 3D coordinates of the target position.

\end{document}